\pdfoutput=1
\documentclass[11pt]{article}

\usepackage[dvipsnames]{xcolor}

\usepackage[]{ACL2023}
\usepackage{times}
\usepackage{latexsym}

\usepackage[T1]{fontenc}
\usepackage[utf8]{inputenc}

\usepackage{microtype}

\usepackage{inconsolata}

\usepackage{booktabs}
\usepackage{graphicx}
\usepackage{multirow}
\usepackage{mdwlist}
\usepackage{amsmath}

\makecompactlist{itemize}{stditemize}

\usepackage{enumitem} %Katie added 

\makeatletter
\newcommand{\ccell}[3][]{%
  \kern-\fboxsep
  \if\relax\detokenize{#1}\relax
    \expandafter\@firstoftwo
  \else
    \expandafter\@secondoftwo
  \fi
  {\colorbox{#2}}%
  {\colorbox[#1]{#2}}%
  {#3}\kern-\fboxsep
}
\makeatother
\definecolor{cellgray}{gray}{0.9}

\title{Words as Gatekeepers: Measuring Discipline-specific \\ Terms and Meanings in Scholarly Publications}

\author{Li Lucy\textsuperscript{1,2} \, Jesse Dodge\textsuperscript{1}\, David Bamman\textsuperscript{2} \, Katherine A. Keith\textsuperscript{1,3} \\ 
\textsuperscript{1}Allen Institute for Artificial Intelligence \\
\textsuperscript{2}University of California, Berkeley \\
\textsuperscript{3}Williams College \\ 
\texttt{\{lucy3\_li, dbamman\}@berkeley.edu} \\
\texttt{jessed@allenai.org} \,
\texttt{kak5@williams.edu}
}

\begin{document}
\maketitle
\begin{abstract}
Scholarly text is often laden with jargon, or specialized language that can facilitate efficient in-group communication within fields but hinder understanding for out-groups. In this work, we develop and validate an interpretable approach for measuring \textit{scholarly jargon} from text. Expanding the scope of prior work which focuses on word types, we use word sense induction to also identify words that are widespread but overloaded with different meanings across fields. We then 
%use normalized pointwise mutual information to 
estimate the prevalence of these discipline-specific words and senses across hundreds of subfields, and show that word senses provide a complementary, yet unique view of jargon alongside word types. We demonstrate the utility of our metrics for science of science and computational sociolinguistics by highlighting two key social implications. First, though most fields reduce their use of jargon when writing for general-purpose venues, and some fields (e.g.,~biological sciences) do so less than others. 
Second, the direction of correlation between jargon and citation rates varies among fields, but jargon is nearly always negatively correlated with interdisciplinary impact. Broadly, our findings suggest that though multidisciplinary venues intend to cater to more general audiences, some fields' writing norms may act as barriers rather than bridges, and thus impede the dispersion of scholarly ideas.
\end{abstract}

% \kkcomment{"bridge --- rather than a barrier --- to broader dispersion of scholarly ideas and findings"}

% \jdcomment{For the abstract, consider starting with something like, ``Scholarly text is often laden with jargon, or specialized language, which emerges naturally in a textual domain. Use of jargon can lead to brevity, while at the same time hindering understanding by those unfamiliar with the relevant context. We explore the use of jargon in scientific text by defining scholarly jargon as discipline-specific word types and sense...''}
% \jdcomment{You don't need to use my suggestion above, but the two main improvements I tried to do are 1. having two sentences at the start introducing what we're working on (jargon is natural, and has positive and negative properties), and 2. i don't think we should start describing our contribution by saying ``we extend past work'', we should start by stating what we did.}
% part 1 is very similar to intro

\section{Introduction}

Specialized terminology, or jargon, naturally evolves in communities as members communicate to convey meaning succinctly. It is especially prevalent in scholarly writing, where researchers use a rich repertoire of lexical choices. However, niche vocabularies can become a barrier between fields \cite{vilhena2014finding,martinez2021specialized,freeling2019how}, and between scientists and the general public \cite{liu2022lexical,august2020explain,cervetti2015factors,freeling2021better}. Identifying scholarly jargon is an initial step for designing resources and tools that can increase the readability and reach of science \cite{august-etal-2022-generating,plaven_sigray2017research,rakedzon2017automatic}.

\begin{figure}[t]
\includegraphics[width=\columnwidth]{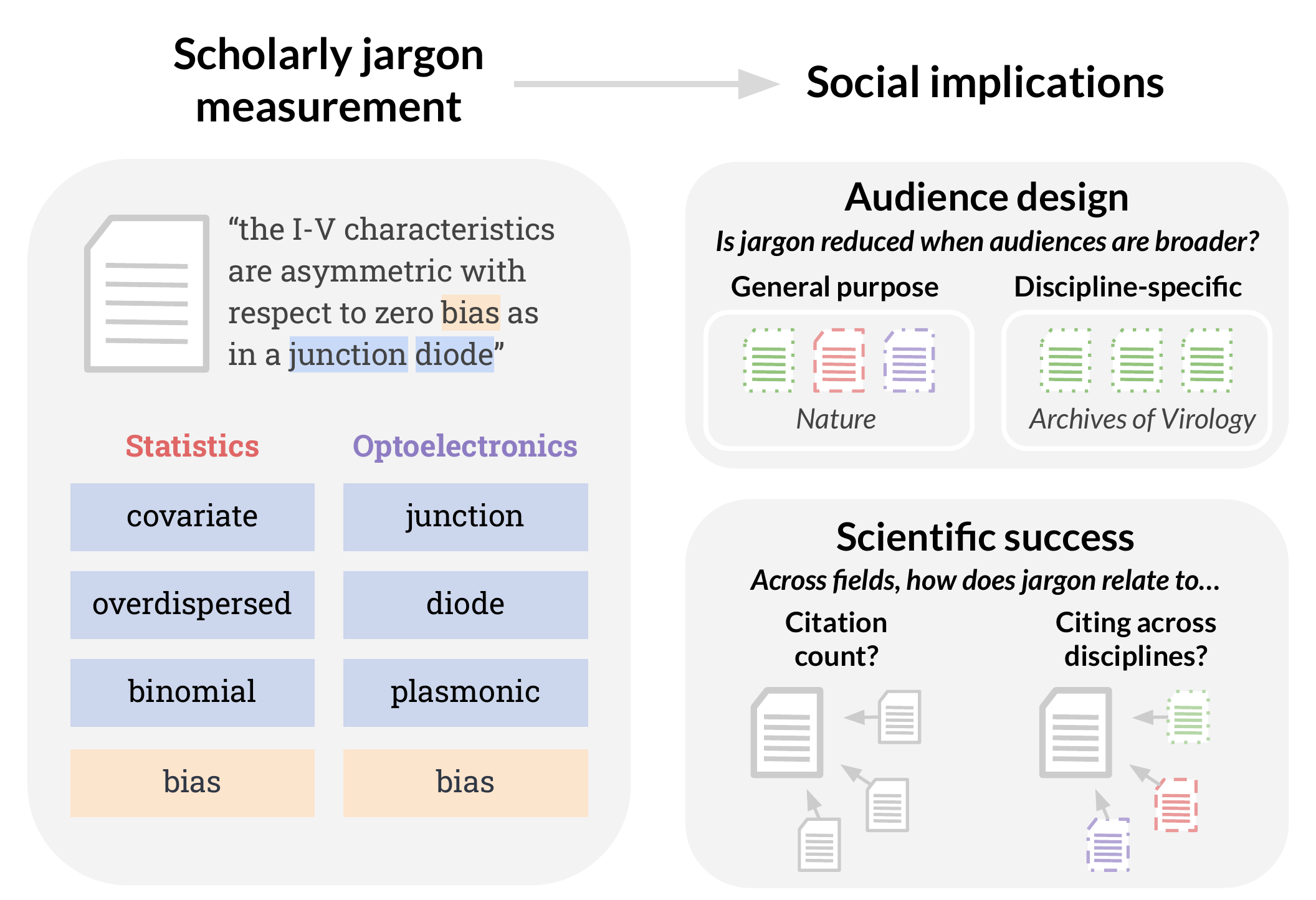}
\centering 
\caption{\label{intro} In this paper, we measure scholarly jargon, which consists of discipline-specific \textcolor{CadetBlue}{word types} and \textcolor{YellowOrange}{senses}. We further illustrate two social implications of jargon: whether its rate differs between broad and narrow audiences (right; top), and how it relates to citation-based success (right; bottom). The example abstract excerpt on the left is from \citet{Satishkumar2000applied}.}
\end{figure}

Research on scholarly language typically focuses on the relative prevalence of words \cite{mckeown2016predicting,prabhakaran-etal-2016-predicting,sim-etal-2012-discovering,rakedzon2017automatic}. However, the same word can be overloaded with multiple meanings, such as \textit{bias} referring to electric currents or statistical misestimation (Figure~\ref{intro}). We use BERT-based word sense induction to disentangle these, and demonstrate the utility of including both word types and senses in our operationalization of \textit{scholarly jargon}. We measure jargon in English abstracts across three hundred fields of study, drawn from over 12 million scholarly abstracts and one of the largest datasets of scholarly documents: the Semantic Scholar Open Research Corpus (S2ORC) \cite{lo-etal-2020-s2orc}.

Our findings are valuable for several groups that partake in science: readers, authors, and science of science researchers. 

% readers
Due to scholarly language's gatekeeping effect, natural language processing (NLP) researchers have developed tools to support \textbf{readers}, such as methods for simplifying or defining terminology \cite{kim-etal-2016-simplescience,vadapalli-etal-2018-science,august-etal-2022-generating,head2021augmenting,august2022paperplain,murthy2022accord}. When deciding what constitutes jargon, studies may rely on vocabulary lists based on word frequency, often collapsing all of science into one homogeneous language variety \cite{august-etal-2020-writing,rakedzon2017automatic,plaven_sigray2017research}. Our approach identifies language associated with individual subfields and proposes a bottom-up, data-driven process for creating these vocabularies (\S\ref{methods} and~\ref{section:val}).

% writers: 
Second, measuring levels of discipline-specific language in abstracts can inform \textbf{authors} who wish to communicate to a wider audience or enter a new field. We show that while some subfields tend to use highly specialized word types, others use highly specialized senses (\S\ref{sec:norms}). In addition, we provide evidence for audience design in scholarly discourse (\S\ref{audience_design}), following a sociolinguistic framework that describes how speakers accommodate language to the scope of their audience \cite{bell_1984}.

% science of science: 
Finally, our language-centered approach contrasts the typical paradigm in \textbf{science of science} research, where citation behavior often defines relationships among articles, venues, and fields \citep[e.g.][]{boyack2005mapping,rosvall2008maps,peng2021neural}. Citation count is a common measurement of ``success'', and the mechanisms behind it form a core research area \cite{wang_barabasi_2021,foster2015tradition,fortunato2018science}. On the other hand, interdisciplinarity is increasingly valued, but does not always lead to short-term citation gains \cite{van2015interdisciplinary,lariviere2010relationship,okamura2019interdisciplinarity,chen2022interdiscip}. We run regression analyses to examine the relationship between discipline-specific senses and types and these two distinct measures of success  (\S\ref{sec:success}). 

% To summarize, our contributions include the following:
To summarize, we contribute the following (Figure~\ref{intro}): 
\begin{itemize}[leftmargin=0.3cm]
  \itemsep0em
    \item \textbf{Methods.} We propose a new measure of scholarly jargon to identify discipline-specific word types and senses (\S\ref{methods}). We validate our approach for measuring senses by showing it recalls more overloaded words in Wiktionary compared to word types alone (Figure~\ref{fig:recall}). 
    \item \textbf{Social implications.} We illustrate the utility of our jargon measurements for computational social science by analyzing audience design and articles' success (\S\ref{social_implic}). Though multidisciplinary venues may intend to be general-purpose, more dominant fields in these venues reduce jargon less so than others (Figure~\ref{fig:audience_design}). Since jargon nearly always has a negative relationship with interdisciplinary impact (Table~\ref{tab:success}), our findings encourage the reconsideration of existing scholarly writing norms. % We find that some fields reduce jargon more than others when authors publish in general-purpose journals  (Figure~\ref{fig:audience_design}), and that jargon has a varying relationship with citation counts, but nearly always a negative one with interdisciplinary impact (Table~\ref{tab:success}). 
    % \kkcomment{I think we're missing the so what, who cares? in both these findings}
\end{itemize}
We hope our measure of scholarly jargon can help researchers quantify language barriers in science and their implications. Our code and scored lists of jargon for each subfield can be found at \href{https://github.com/lucy3/words_as_gatekeepers}{https://github.com/lucy3/words\_as\_gatekeepers}.

\section{Data}

Our work involves several datasets: scholarly abstracts, Wikipedia, and Wiktionary. We use abstracts to calculate the association of words with disciplines and Wikipedia to supplement our calculation of background word probabilities. Later, in \S\ref{section:val}, we introduce and describe how we use Wiktionary to validate our approach. 

\begin{figure}[t]
    \includegraphics[width=\columnwidth]{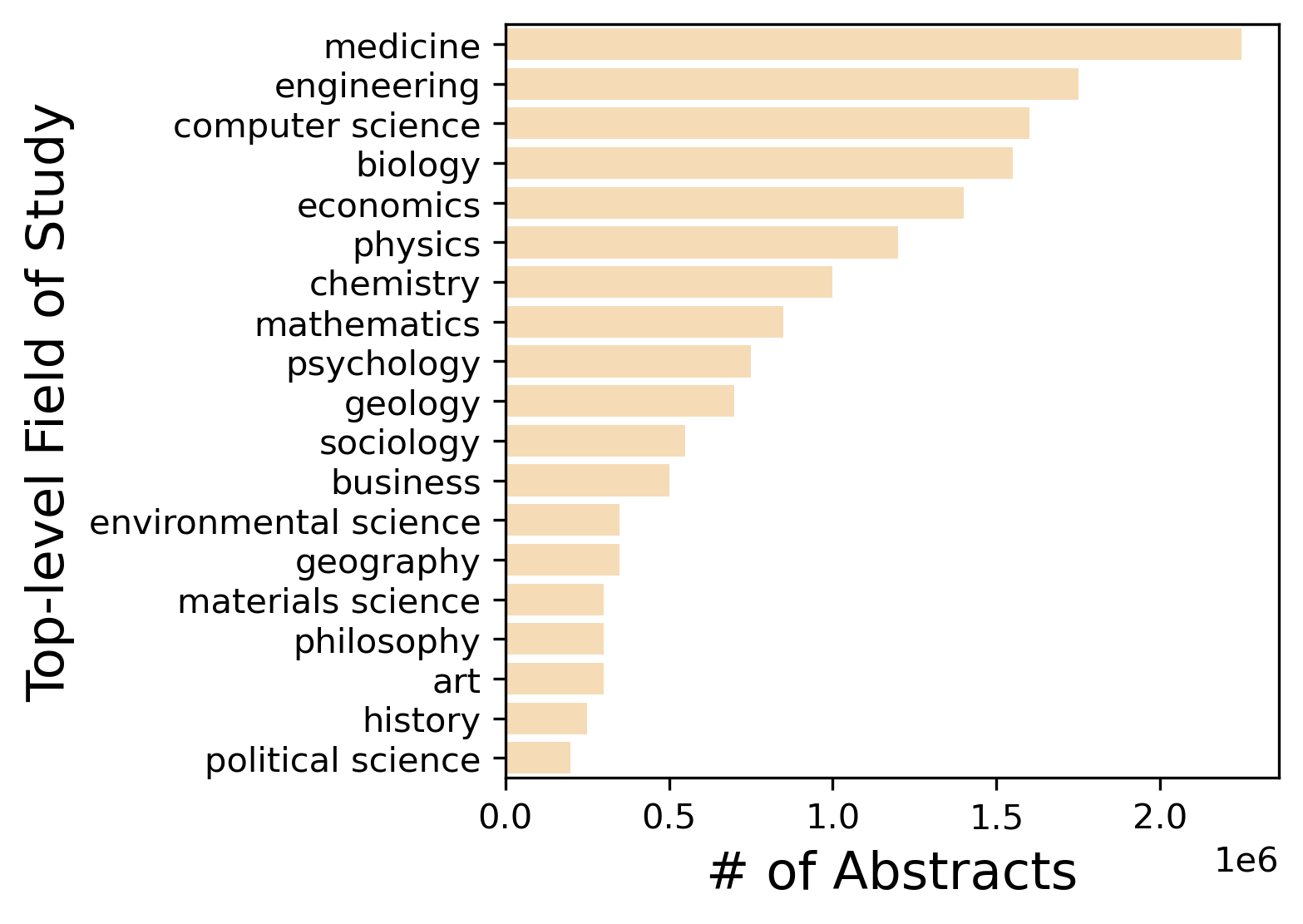}
    \centering
    \caption{The number of abstracts in each top-level scholarly field in \textsc{contemporary s2orc}, sorted by size. In total, there are 12.0 million abstracts.}
	\label{fig:fos_data}
\end{figure}

\subsection{Contemporary S2ORC}

Our dataset of academic articles, \textsc{contemporary s2orc}, contains 12.0 million abstracts and 2.0 billion words\footnote{We define a word as a non-numeric, non-punctuation token outputted by Huggingface transformer's whole-word BasicTokenizer.} that span a mix of scholarly fields (Figure~\ref{fig:fos_data}, Appendix~\ref{appx:fos}).

To create \textsc{Contemporary S2ORC}, we draw from the July 2020 release of S2ORC \cite{lo-etal-2020-s2orc}. S2ORC is a general purpose corpus that contains metadata for 136 million scholarly articles, including 380.5 million citation links \cite{lo-etal-2020-s2orc}. These articles originate from Semantic Scholar, which obtains data directly from publishers, the Microsoft Academic Graph (MAG), arXiv, PubMed, and the open internet. Metadata, such as titles, authors, publication years, journals/venues, and abstracts are extracted from PDFs and LaTeX sources or provided by the publisher. Though extensive, S2ORC contains some amount of noisy or missing metadata. We remove non-English articles and those with missing metadata, consolidate journals and venues into a single venue label, and limit the dataset to the years 2000-2019 (Appendix~\ref{appx:filtering}).

S2ORC links articles to paper IDs in the Microsoft Academic Graph (MAG) \cite{sinha2015mag,wang2019mag}, so we match S2ORC abstracts to MAG fields of study (FOS). S2ORC originally contains top-level MAG FOS (level 0), e.g. \textit{biology}, but we also join abstracts with second level MAG FOS (level 1), e.g. \textit{immunology}, for more granularity.\footnote{A secondary level FOS may fall under multiple top-level FOS, some articles are labeled with multiple FOS at the same level, and some articles marked with top-level FOS do not indicate a secondary level FOS.} In this present paper, we refer to level 0 FOS as \textit{fields}, and level 1 FOS as \textit{subfields}. We take an approximately uniform sample of 50k articles per subfield, resulting in a total of 293 subfields that fall under 19 fields (Appendix~\ref{appx:fos}). 

\subsection{Wikipedia}

\begin{table*}[t]
\centering
\resizebox{\textwidth}{!}{%
\begin{tabular}{@{}cccccccccccc@{}}
\toprule
\multicolumn{2}{c}{NLP} & \multicolumn{2}{c}{Chemical Engineering} & \multicolumn{2}{c}{Immunology} & \multicolumn{2}{c}{Communication} & \multicolumn{2}{c}{International Trade} & \multicolumn{2}{c}{Epistemology}\\
\cmidrule(lr){1-2}\cmidrule(lr){3-4}\cmidrule(lr){5-6}\cmidrule(lr){7-8}\cmidrule(lr){9-10}\cmidrule(lr){11-12}
word  & $\mathcal{T}_f(t)$  & word & $\mathcal{T}_f(t)$  & word & $\mathcal{T}_f(t)$  & word & $\mathcal{T}_f(t)$  & word & $\mathcal{T}_f(t)$  & word & $\mathcal{T}_f(t)$\\
\midrule
nlp & 0.412 & rgo & 0.334 & treg & 0.346 & saccade & 0.354 &wto& 0.453&epistemic&0.356\\ 
corpora & 0.404 & mesoporous & 0.328 & cd4 & 0.341 & saccades& 0.345 &trade&0.438 &epistemology&0.350\\ 
treebank & 0.401 & nanosheets & 0.327 & immune & 0.3388 & stimuli& 0.333&fdi& 0.401&epistemological&0.342\\  
disambiguation & 0.396 & nanocomposite & 0.325 & il & 0.336 & stimulus&0.331 &ftas&0.396&husserl&0.332\\ 
corpus & 0.393 & nanocomposites & 0.324 & th2 & 0.335 & cues& 0.327&antidumping&0.396 &kant&0.329\\ 
\bottomrule
\end{tabular}%
}
\caption{Top five words that are highly specialized to different disciplines. These have the highest type NPMI ($\mathcal{T}_f(t)$) scores in their respective subfields. As examples, \textit{treg} in immunology stands for ``regulatory T cells'', and \textit{antidumping} in international trade places high taxes on imports.}
\label{tab:top_type_examples}
\end{table*}

We include Wikipedia article content to counterbalance \textsc{Contemporary S2ORC}'s STEM-heavy focus for our estimation of words' typical prevalence. Wikipedia is a popular information-gathering resource \cite{reagle2020wikipedia}, and we use an Oct 1, 2022 dump of its articles. It offers complementary topical coverage that is collectively curated and driven by public interest, and includes biographies, culture, and arts \cite{mesgari2015sum}. We remove Wiki formatting using \citet{Wikiextractor2015}'s text extractor, and discard all lines, or paragraphs, that are less than 10 white-spaced tokens long. We sample twice as many Wikipedia paragraphs as the number of \textsc{Contemporary S2ORC} abstracts, so that each is similar in size despite differences in document length. In total our Wikipedia dataset, \textsc{WikiSample}, contains $24.0$ million paragraphs and 1.5 billion tokens. 

\section{Methods}\label{methods}

Language differences among subsets of data can be measured by a variety of approaches, from geometric to information theoretic \cite{ramesh-kashyap-etal-2021-domain,vilhena2014finding,aharoni-goldberg-2020-unsupervised}. We calculate the association of a word's type or sense to subfields using normalized pointwise mutual information (NPMI). We choose NPMI over similar metrics (e.g. tf-idf, divergence, $z$-score) because of the nature of language difference it emphasizes: higher NPMI scores reflect language that is not only commonly used in a community, but also highly specific to it \cite{lucy-bamman-2021-characterizing,gardner-etal-2021-competency}. NPMI offers an interpretable metric of association, where a score of 1 indicates perfect association, 0 indicates independence, and -1 indicates no association. We follow \citet{lucy-bamman-2021-characterizing}'s framework of calculating NPMI separately for word types and senses, which they originally used to identify community-specific language on social media. We update their approach with a more recent word sense induction (WSI) method, and use a different interpretation of type and sense NPMI scores.

\subsection{Discipline-specific words}

We calculate NPMI for word types, or \textit{type NPMI}, as the following measure::
\begin{equation}
\mathcal{T}_f(t) = \frac{\log{ (P(t \mid f) / P(t))}}{-\log{P(t, f)}}.
\end{equation}
Here, $P(t \mid f)$ is the probability of a word $t$ occurring given a set of abstracts $f$ in a field, $P(t, f)$ is their joint probability, and $P(t)$ is the probability of the word overall \cite{lucy-bamman-2021-characterizing,Zhang2017community}. ``Overall'' refers to the combined background dataset of \textsc{Contemporary S2ORC} and \textsc{WikiSample}. We only calculate $\mathcal{T}_f(t)$ for words that appear at least 20 times in each field.\footnote{As we will describe in \S\ref{method:senses}, the sense NPMI pipeline operates on lemmas, not words. Standard lemmatizers may not be suitable for rarer words in science, so to make our type and sense metrics comparable, we only lemmatize the set of widely-used words that are shared by both pipelines.}

As illustrative examples, Table~\ref{tab:top_type_examples} shows words with the highest $\mathcal{T}_f(t)$ in several fields. 

\subsection{Discipline-specific senses}\label{method:senses}

Widely disseminated words can be overloaded with domain-specific meanings or use. For example, \textit{bias} could refer to a type of voltage applied to an electronic system, social prejudice, or statistical misestimation. Thus, we include word senses as a complement to word types for characterizing domain-specific language. We use \textit{senses} to refer to different meanings or uses of the same word induced by word sense induction (WSI).

\subsubsection{Word sense induction}

To partition occurrences of words into senses, we adapt \citet{eyal-etal-2022-large}'s WSI pipeline with minimal modifications. WSI is an unsupervised task where occurrences of words are split into senses. \citet{eyal-etal-2022-large}'s approach is designed for large-scale datasets, where a sample of a target word's occurrences is used to induce senses, and remaining occurrences are then assigned to them. To induce senses, a masked language model predicts the top $s$ substitutes of each occurrence of a target word. Then, a network is created for each target word, where nodes are substitutes and edges are their co-occurrence. Louvain community detection is then applied to determine senses, or sets of substitutes \cite{Blondel_2008}. For example, in the network for \textit{bass}, the substitutes for its sense as a type of fish are likely not predicted at the same time as substitutes for its musical sense, so each set would represent separate senses. 

We carry out this WSI pipeline on a case-insensitive target vocabulary of 6,497 ``widely used'' words: those that appear in the top 98th percentile by frequency and in at least 50\% of venues, not including stopwords and words split into wordpieces.\footnote{We avoid wordpieces since \citet{eyal-etal-2022-large}'s pipeline predicts substitutes at the token-level.} We lemmatize and lowercase target words and substitutes, following \citet{eyal-etal-2022-large}'s implementation, because otherwise the most common substitutes representing a sense may be different lemmas of the same word. This processing step reduces the target vocabulary into 4,407 lemmas. 

We sample 1000 instances of each vocabulary lemma, and use ScholarBERT to predict each instance's top $s=5$ substitutes \cite{scholarBERT2022}.\footnote{We pick ScholarBERT over similar transformer models trained on scholarly language (e.g.~SciBERT), because it is trained on a wider breadth of disciplines, splits fewer potential vocab words into wordpieces (15 versus 193), and uses RoBERTa-style training \cite{scholarBERT2022,beltagy-etal-2019-scibert}.} We truncate each abstract to this model's maximum input length. We follow \citet{eyal-etal-2022-large}'s heuristics for determining sets of substitutes that are big enough to recognize as senses: each set needs to have at least two substitutes, and the second most frequent substitute needs to appear at least 10 times across the target word's sample. If no sets are big enough, we add a fallback case, where we place all occurrences of a word to a single sense. 

\citet{eyal-etal-2022-large} assigns additional occurrences of the target word to induced senses based on Jaccard similarity. We also add a fallback case here: if the overlap of a remaining occurrence's substitutes with all senses is zero, we assign that occurrence to an extra sense representing previously unseen senses. 

\subsubsection{Sense NPMI}

Once each occurrence of a widely-used word is labeled with a sense, their frequencies can be used to calculate sense NPMI. Sense NPMI uses the same formula as type NPMI, except it is calculated at the sense-level rather than the word-level \cite{lucy-bamman-2021-characterizing}. That is, counts of a word $t$ are replace with counts of its $i$th sense, $t_i$: 
\begin{equation}
\mathcal{S}_f(t_i) = \frac{\log{ (P(t_i \mid f) / P(t_i))}}{-\log{P(t_i, f)}}.
\end{equation}

\section{Validation}\label{section:val}

\subsection{Wiktionary}

We perform in-domain validation of the unsupervised sense pipeline using Wiktionary. Words marked as associated with a subfield in this online dictionary should also be highly scored by our metrics. Wiktionary is collaboratively maintained and includes common words listed with definitions that may be labeled as having ``restricted usage'' to a topic or context. For example, the word \textit{ensemble} has the labels \textit{machine learning}, \textit{fashion}, and \textit{music} (Appendix~\ref{appx:val}). We map Wiktionary labels in English definitions of target words using exact string matching to fields and subfields. If an NPMI score threshold were used to determine whether a token should be considered discipline-specific or not, we expect sense NPMI to recall more words labeled by Wikitionary than type NPMI does. We do not calculate precision, because Wiktionary is not necessarily comprehensive for all subfields.

We obtain Wiktionary entries for 94.94\% of the common, widely used words that were inputs in the WSI pipeline. We filter out words where all definitions are labeled with only one field, and allow subfields to inherit the words labeled with their parent field. In total, we have 11,548 vocabulary word and subfield pairs to recall across 83 subfields. Since recall is calculated at the word-level and sense NPMI is at the sense-level, we use a word $t$'s most frequent sense $t_i$'s $\mathcal{S}_f(t_i)$ in a subfield to represent word-level sense NPMI $\mathcal{S}_f(t)$.

\begin{figure}[t]
\begin{minipage}{0.5\columnwidth}
\includegraphics[width=.98\linewidth]{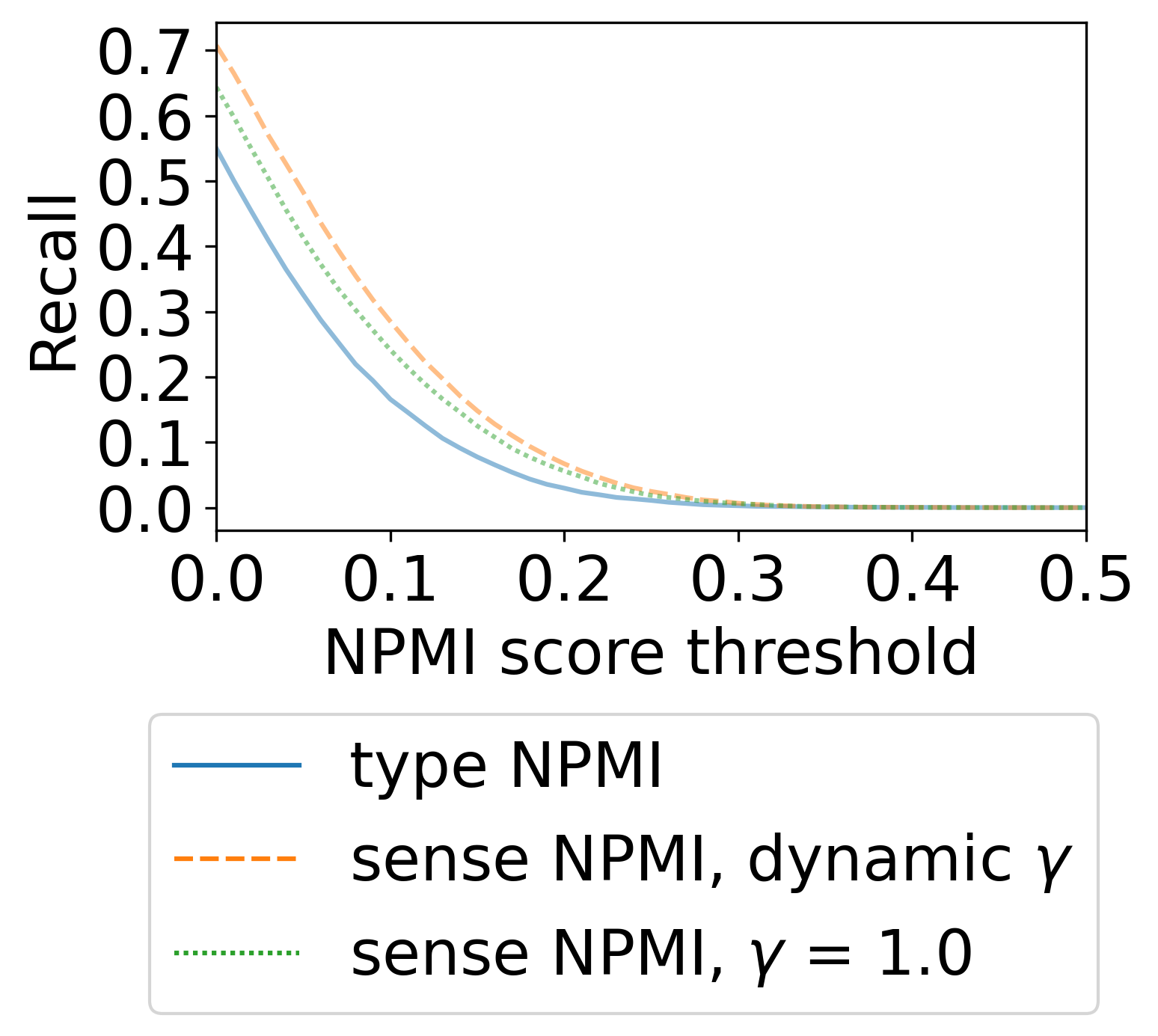}
\end{minipage} \hfill
\begin{minipage}{0.4\columnwidth}
\resizebox{0.98\linewidth}{!}{%
    \begin{tabular}{@{}lc@{}}
\toprule
NPMI metric & AUC, recall \\ \midrule
$\mathcal{S}_f(t)$, $\gamma = 0.5$ & 0.0550 \\
$\mathcal{S}_f(t)$, $\gamma = 1.0$ & 0.0583 \\
$\mathcal{S}_f(t)$, $\gamma = 1.5$ & 0.0631 \\
$\mathcal{S}_f(t)$, $\gamma = 2.0$ & 0.0670 \\
$\mathcal{S}_f(t)$, $\gamma = 2.5$ & 0.0697 \\
$\mathcal{S}_f(t)$, dynamic $\gamma$ & 0.0675 \\
\midrule
$\mathcal{T}_f(t)$ baseline  & 0.0434  \\ \bottomrule
\end{tabular}%
}
\end{minipage}
    % \centering
    \caption{Recall and area under the curve (AUC) of 11,548 Wiktionary words with discipline-specific definitions. Sense NPMI with dynamic resolution ($\gamma$) recalls more semantically overloaded words than type NPMI at the same score threshold. 
    \label{fig:recall}}
	
\end{figure}

In \citet{eyal-etal-2022-large}'s WSI pipeline, the resolution parameter $\gamma$ in Louvain community detection calibrates the number of senses induced per word. Increasing resolution leads to more fine-grained word senses and higher recall, but potentially spurious senses (Figure~\ref{fig:recall}). Rather than using \citet{eyal-etal-2022-large}'s default resolution value of 1, we use a dynamic formula for resolution \cite{newman2016equivalence}: 
\begin{equation}
\gamma = \frac{\omega_{in} - \omega_{out}}{\log \omega_{in} - \log \omega_{out}},
\end{equation}
where $\omega_{in}$ is the probability of an edge between two nodes in the same community, and $\omega_{out}$ is the probability of an edge between two nodes in different communities. Intuitively, nodes within communities should be more connected than nodes between them. We follow \citet{newman2016equivalence}'s algorithm, initializing $\gamma = 1$ and iterating for each target lemma at most 10 times. In each iteration, we run Louvain community detection and recalculate $\gamma$ using the edge probabilities in the current clustering. We stop early if $\gamma$ converges within 0.01 of its previous value. 

Sense NPMI with dynamic resolution recalls more discipline-specific Wiktionary words than type NPMI at the same score cutoff (Figure~\ref{fig:recall}). In addition, the sense NPMI of a word in a subfield labeled by Wiktionary is higher than the score of the same word in a random field (paired $t$-test, $p < 0.001$, Appendix~\ref{appx:val}). Thus, Wiktionary-based validation shows that our unsupervised approach is able to measure discipline-specific senses, and in all downstream analyses, we use the dynamically defined $\gamma$ for WSI. 

\begin{table*}[t]
\centering
\resizebox{\textwidth}{!}{%
\begin{tabular}{@{}ccccccc@{}}
\toprule
          & \multicolumn{3}{c}{sense $t_1$}            & \multicolumn{3}{c}{sense $t_2$}             \\
\cmidrule(lr){2-4} \cmidrule(lr){5-7} 
word $t$ & FOS $a$    & $\mathcal{S}_a(t_1)$ & top substitutes & FOS $b$     & $\mathcal{S}_b(t_2)$ & top substitutes \\ \midrule
\textit{kernel} & Operating system & 0.321   & block, personal, ghost, every, pure & Agronomy & 0.272 & grain, palm, body, gross, cell \\
\textit{performance} & Chromatography & 0.266  & perform, play, timing, temperature, contribute
 & Industrial organization & 0.234 & success, record, position, accomplishment, hand \\
\textit{network} & Computer network & 0.327  & graph, net, regular, key, filter
& Telecommunications & 0.259 & connection, channel, link, connectivity, association \\
\textit{root} & Dentistry & 0.413 & crown, arch, tooth, long, tissue & Horticulture & 0.330 & plant, tree, branch, part, stem \\
\textit{power} & Electrical engineering & 0.329  & energy, electricity, load, fuel, lit        & Combinatorics & 0.193 & value, order, term, sum, degree \\
\bottomrule
\end{tabular}%
}
\caption{Hand-selected words that are common across fields, but have different uses or meanings. The senses shown for each word are the two with the highest sense NPMI scores for that word across fields. Each sense is represented by the five most common substitutes suggested by ScholarBERT for instances in that sense.}
\label{tab:sense_examples}
\end{table*}

\subsection{Examples and interpretation}

Examples of semantically overloaded words between fields can also lend face validity to our results (Table~\ref{tab:sense_examples}). Returning to the example introduced at the beginning, \textit{bias} is indeed very overloaded. It has distinct senses with high NPMI (> 0.2) across multiple fields, including statistics (\textit{skew}),\footnote{Word in parentheses is the top predicted substitute for that subfield's sense for \textit{bias}.} optoelectronics (\textit{charge}), cognitive psychology (\textit{preference}), and climatology (\textit{error}). These examples suggest that future work could examine how our approach could provide potential candidates for updating dictionaries or glossaries when new senses are introduced.

\begin{table*}[t]
\centering
\resizebox{\textwidth}{!}{%
\begin{tabular}{@{}cccc cccc cccc cccc@{}}
\toprule
\multicolumn{4}{c}{Pure mathematics} & \multicolumn{4}{c}{Monetary economics} & \multicolumn{4}{c}{Computer security} & \multicolumn{4}{c}{Stereochemistry} \\
\cmidrule(lr){1-4}\cmidrule(lr){5-8} \cmidrule(lr){9-12}\cmidrule(lr){13-16}
word & $\Delta$ & $\mathcal{S}_f(t)$ & $\mathcal{T}_f(t)$ & word & $\Delta$ & $\mathcal{S}_f(t)$ & $\mathcal{T}_f(t)$ & word & $\Delta$ & $\mathcal{S}_f(t)$ & $\mathcal{T}_f(t)$ & word & $\Delta$ & $\mathcal{S}_f(t)$ & $\mathcal{T}_f(t)$\\ \midrule
power & 0.202 & 0.186 & -0.016 & movement & 0.218 & 0.266 & 0.048 & primitive & 0.162 & 0.221 & 0.058 & attack & 0.228 & 0.184 & -0.044 \\
pole & 0.194 & 0.207 & 0.013 & liquid & 0.195 & 0.196 & 0.002 & host & 0.151 & 0.205 & 0.054 & title & 0.216 & 0.264 & 0.048 \\
union & 0.193 & 0.141 & -0.051 & interest & 0.182 & 0.382 & 0.200 & elasticity & 0.148 & 0.158 & 0.010 & km & 0.212 & 0.175 & -0.037\\
surface & 0.193 & 0.260 & 0.068 & turbulence & 0.176 & 0.155 & -0.021 & hole & 0.147 & 0.134 & -0.013 & framework & 0.205 & 0.215 & 0.010\\
origin & 0.193 & 0.188 & -0.005 & provider & 0.176 & 0.121 & -0.055 & key & 0.142 & 0.320 & 0.179 & solve & 0.202 & 0.165 & -0.037\\
\bottomrule
\end{tabular}%
}
\caption{Top five words that have senses associated with each subfield ($\mathcal{S}_f(t) > 0.1$), ordered by the difference $\Delta$ between word-level sense and type NPMI. These are words that are highly specific to subfields based on their sense, rather than their type. As examples, monetary economics uses \textit{liquid} to describe valuables that can be easily converted to cash, and stereochemistry uses \textit{attack} to refer to the addition of atoms or molecules during chemical reactions.}
\label{tab:top_sense_examples}
\end{table*}

Table~\ref{tab:top_sense_examples} shows examples of words whose scores increase from type NPMI to sense NPMI despite having counts split across senses. \citet{lucy-bamman-2021-characterizing} interpret sense and type NPMI similarly in their downstream analyses, based on the magnitude of their values, but this does not account for how type and sense NPMI scores are related. In the boundary case where a word $t$ only has a single sense $t_0$, $\mathcal{S}_f(t_0) = \mathcal{T}_f(t)$. This leads to a strong correlation between the two metrics, especially when a sense scored as highly associated with a field is also the dominant sense of that word in general. Thus, to narrow in on what we gain from WSI, we examine not only senses that are highly associated with a field, but have sense NPMI scores higher than their words' type NPMI scores (Table~\ref{tab:top_sense_examples}). Therefore, we count a token with a labeled sense as 
a \textit{discipline-specific sense} if $\mathcal{S}_f(t_i) > \mathcal{T}_f(t)$ and $\mathcal{S}_f(t_i) > c$ for a subfield $f$ and some cutoff $c$. Otherwise, the token is a \textit{discipline-specific type} if $\mathcal{T}_f(t) > c$. 

\section{Language norms across fields}\label{sec:norms}

The linguistic insularity of science varies across fields. For example, \citet{vilhena2014finding} found that phrase-level jargon separates biological sciences more so than behavioral and social sciences. We perform a similar analysis with the novel addition of word senses. 

To summarize the distinctiveness of word types in a field, we calculate the mean type NPMI score of unique words in a field. Before taking the mean, however, we adjust scores by zeroing negative values, since we are more interested in words associated with a field rather than those that are not. This zeroing practice is typically used in studies where PMI measures word relatedness \cite{levy-etal-2015-improving, dagan-etal-1993-contextual,bullinaria2007extracting}.

\begin{figure}[t]
    \includegraphics[width=\columnwidth]{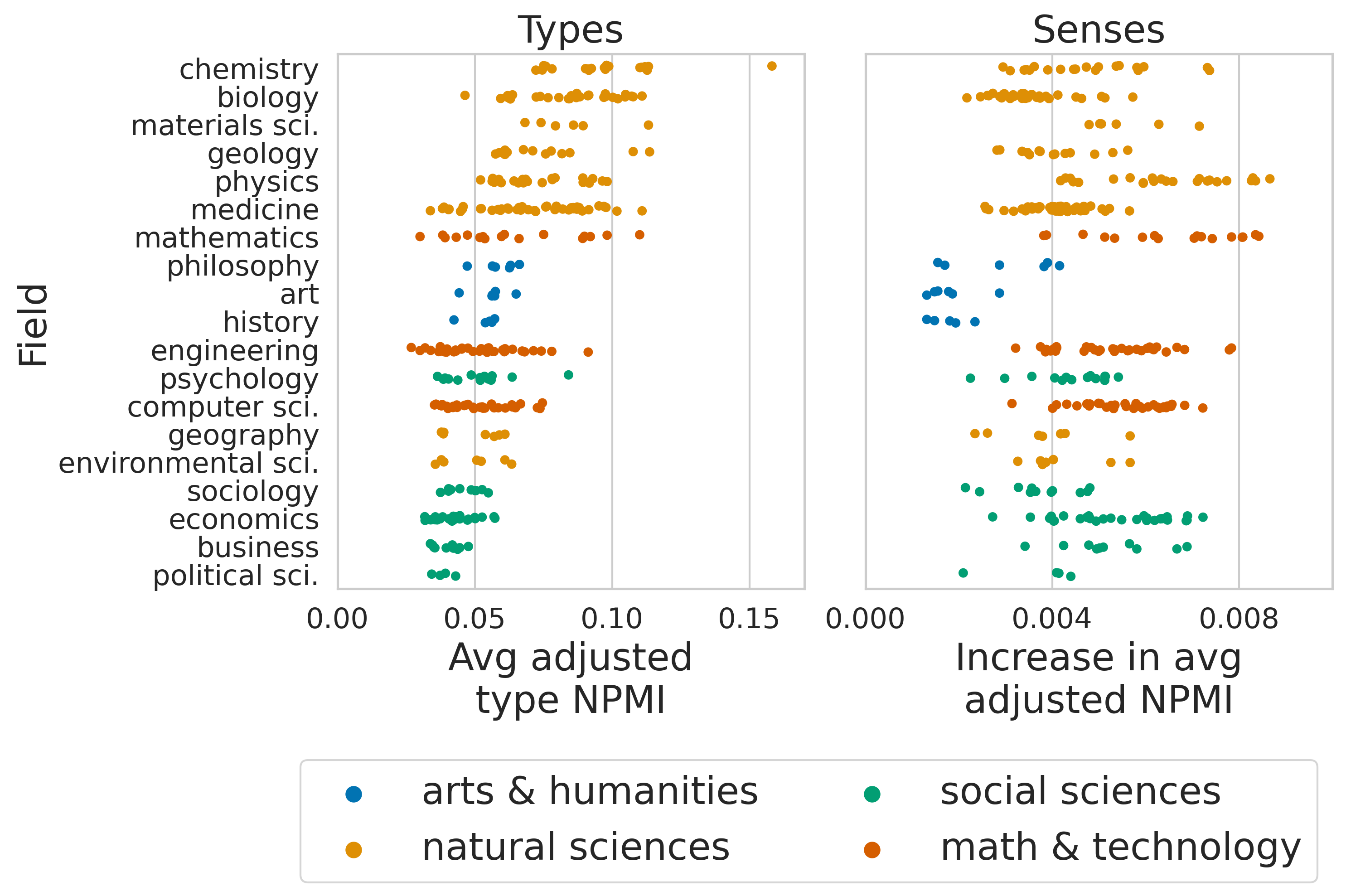}
    \centering
    \caption{The left subplot shows the distinctiveness of subfields' word types, while the right shows the increase in distinctiveness when we take the max of words' type and sense NPMI instead of only their type NPMI. Each point is a subfield, such as \textit{organic chemistry} in \textit{chemistry}, and fields are colored using larger disciplinary categories for interpretation clarity.}
	\label{fig:level_1_jargon}
\end{figure}

Like \citet{vilhena2014finding}, we also find that the biological sciences have very distinctive word types (Figure~\ref{fig:level_1_jargon}). However, there is a considerable amount of overlap in word type distinctiveness across fields. Similar to how natural sciences name molecules and chemicals, the arts and humanities name canons of writers, philosophers, and artists. 

We also examine what fields gain the most in NPMI scores when common words are broken into their senses. We recalculate subfields' average adjusted NPMI, but use $\max(\mathcal{T}_f(t), \mathcal{S}_f(t))$ instead of $\mathcal{T}_f(t)$ for words that have induced senses. Based on their relative increases in average adjusted NPMI, subfields in math/technology, physics, and economics often use common words in specialized contexts (Table~\ref{tab:top_sense_examples}, Figure~\ref{fig:level_1_jargon}). There is no significant Pearson correlation between the distinctiveness of subfields' word types and that of their senses. Thus, word senses provide a very different perspective on language norms and suggests an additional route through which gatekeeping may occur. 

\section{Social implications}\label{social_implic}

In this section, we examine two social implications of our metrics: audience design and scholarly success. We limit these experiments to articles in \textsc{Contemporary S2ORC} that are published among 11,047 venues in the top 95th percentile by abstract count (at least 800 each in S2ORC), to ensure solid estimation of venue-level information, such as their disciplinary focus and average citations per article. 

\subsection{Audience design}\label{audience_design}

Audience design is a well-studied sociolinguistic phenomenon where a speaker's language style varies across audiences \cite{bell_1984,bell_2002,ndubuisi-obi-etal-2019-wetin,androutsopoulos2014languaging}. For example, on Twitter, when writers target smaller or more geographically proximate audiences, their use of nonstandard language increases \cite{pavalanathan2015audience}. Here, we examine this type of language accommodation at the level of subfields, as our data does not contain unique author identifiers that would allow measurements of author-level variation. We hypothesize that for abstracts within the same subfield, ones published for broader audiences (general-purpose venues) use less scholarly jargon than those published in narrower, discipline-focused venues. 

To address this hypothesis, we first collect sets of 6 general-purpose and 2464 discipline-specific venues. We use general-purpose venues that appear in both our dataset and Wikipedia's list of general and multidisciplinary journals:\footnote{\href{https://en.wikipedia.org/wiki/List_of_scholarly_journals}{https://en.wikipedia.org/wiki/List\_of\_scholarly\_journals}.} \textit{Nature}, \textit{Nature Communications}, \textit{PLOS One}, \textit{Science}, \textit{Science Advances}, and \textit{Scientific Reports}. Discipline-focused venues are those where 80\% of articles fall under a single subfield or its name contains the subfield, e.g. \textit{\underline{Agronomy} Journal}.\footnote{There is substantial overlap between these two groups, where 80\% of venues dominated by one subfield also mention the subfield in its name.} Among these two venue sets, we examine abstracts labeled with only one subfield. 

\begin{figure}[t]
    \includegraphics[width=\columnwidth]{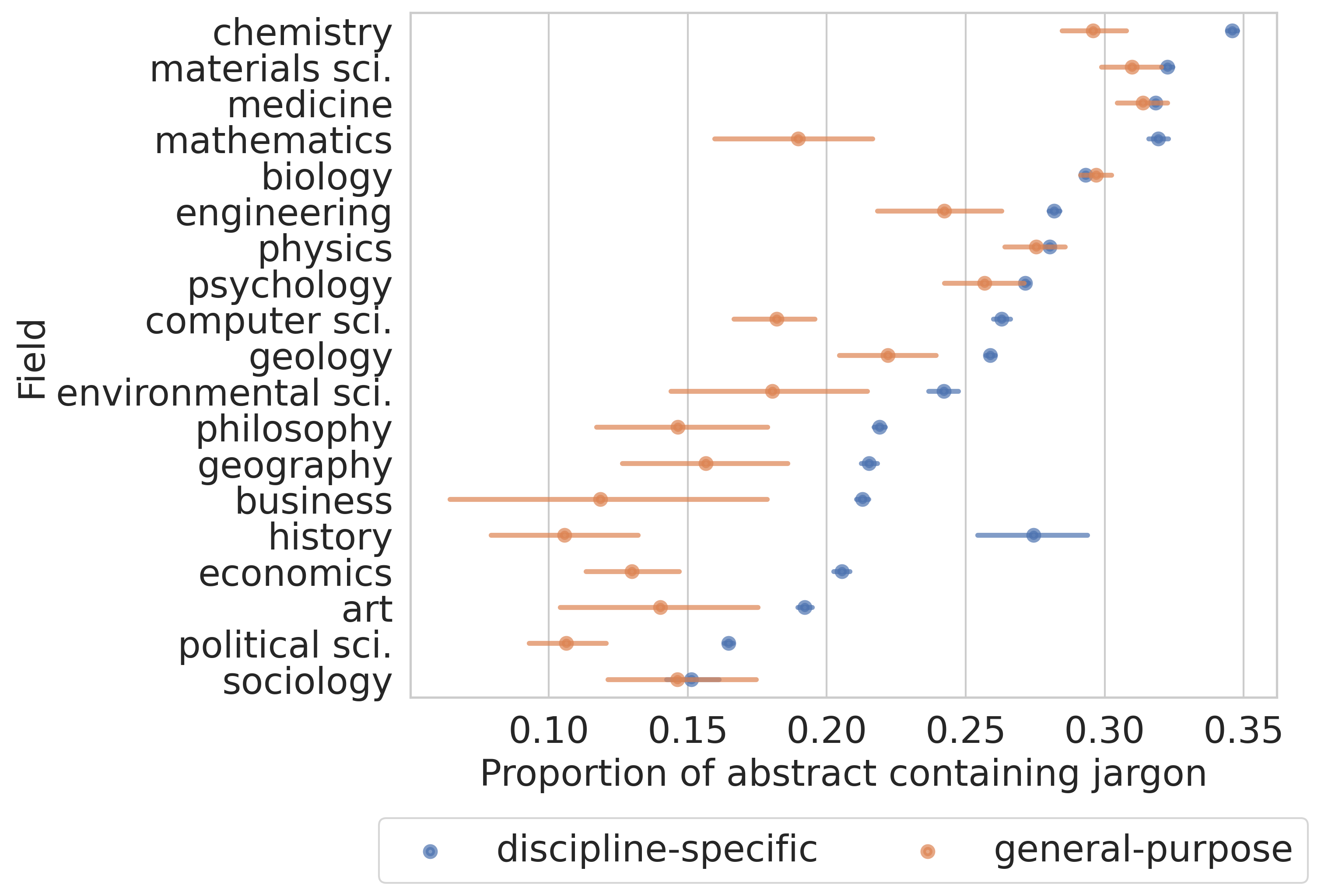}
    \centering
    \caption{Abstracts in the same field typically contain less jargon when they appear general-purpose venues (e.g. \textit{Nature}) than when they are in discipline-focused ones (e.g. \textit{Genetics}). Fields are ordered by monotonically decreasing averages, and error bars are 95\% CI.}
	\label{fig:audience_design}
\end{figure}

We then calculate the fraction of jargon over all words in each abstract, by counting tokens $t$ that are either discipline-specific senses or types, where $c = 0.1$. In other words, we count $t$ if $\max(\mathcal{T}_f(t), \mathcal{S}_f(t)) > 0.1$ in the abstract's subfield $f$. We find that most fields adjust their rate of jargon based on audience, though fields such as medicine and physics are notable exceptions (Figure~\ref{fig:audience_design}). One explanation for this exceptional behavior is that general-purpose venues have a history of being led and dominated by biological sciences, and in some, by physical sciences as well \cite{de2017multidisciplinarity,koopman2011scientific,varmus2000open}. Thus, jargon-laden fields further from these areas adjust their writing the most when publishing in these venues.

\begin{figure}[t]
    \includegraphics[width=\columnwidth]{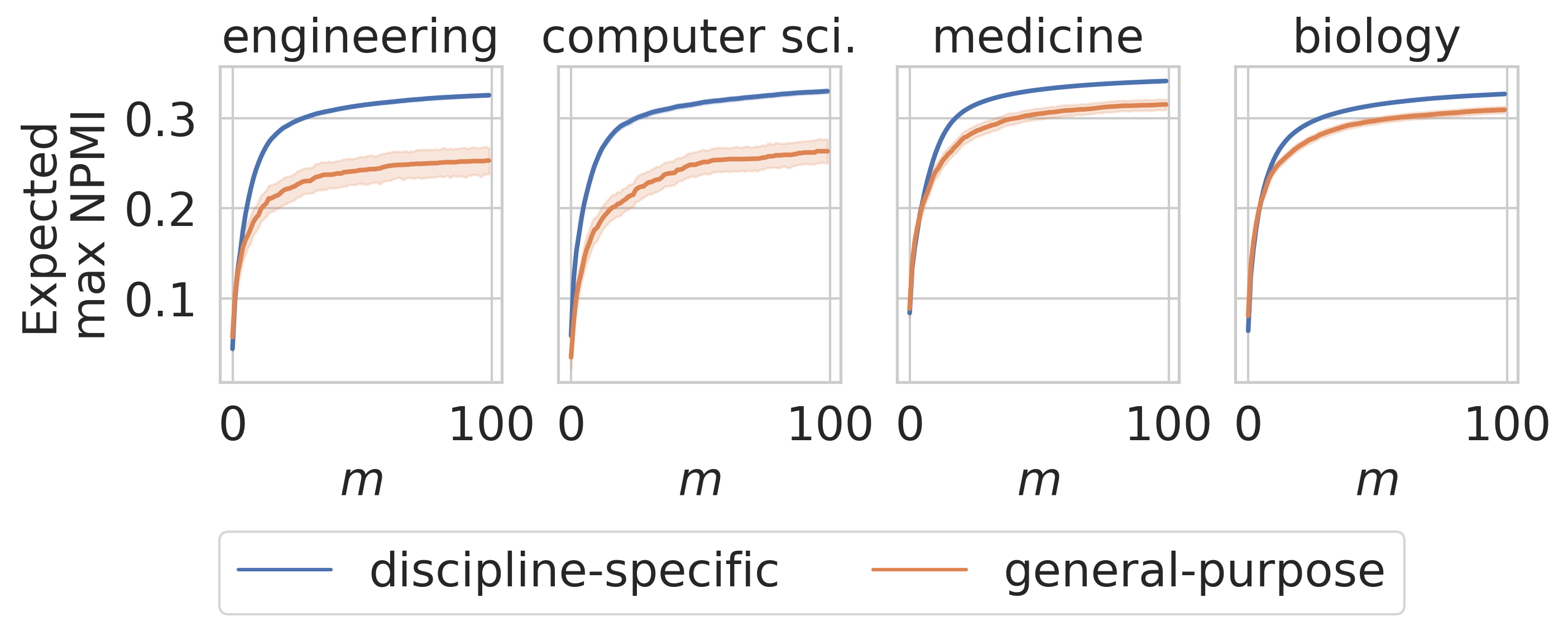}
    \centering
    \caption{The expected maximum type and sense NPMI of tokens in abstracts, where $m$ on the $x$-axis indicates token position and shaded areas are 95\% CI. Computer science and engineering have wider gaps in scholarly jargon use between general-purpose and discipline-focused venues than medicine and biology do.}
	\label{fig:max_expected_npmi}
\end{figure}

A limitation of this approach for quantifying the amount of jargon in an abstract is that it relies on choosing $c$. We also obtain similar results with $c = 0.2$ and justify our choice of $c$ in Appendix~\ref{appx:audience}. An alternative perspective mimics how soon a reader may encounter highly specialized language in an abstract. In this approach, we calculate the maximum over an abstract's type or sense NPMI scores within the first $m$ tokens of the abstract. These results provide another view of our previous finding: fields such as computer science and engineering adjust their content for general-purpose venues more so than those in the biological sciences (Figure~\ref{fig:max_expected_npmi}). This indicates that though most ``general-purpose'' venues intend to be for all of science,\footnote{For example, \textit{Nature}'s ``Aims and Scope'': ~\href{https://www.nature.com/nature/journal-information}{https://www.nature.com/nature/journal-information}.} some fields are expected to adapt their language more so than others. 

Among in-group members, the use of specialized vocabulary can signal legitimacy and expertise \cite{agha2005registers,labov1973sociolinguistic}. Thus, there may be competing incentives influencing authors' writing. In the next section, we further investigate the relationship between jargon and two incentives in science: citation count and a metric of interdisciplinary impact.

\subsection{Scholarly success}\label{sec:success}

We hypothesize that jargon plays different roles in the success of an article depending on how ``success'' is defined. In particular, since jargon gatekeeps outsiders from a discipline, we expect it to negatively affect interdisciplinary impact. To test this hypothesis, we run two sets of regressions to measure the relationship between abstracts' use of jargon and citation behavior within five years after publication. The first set of regressions predicts short-term citation counts, while the second predicts interdisciplinary impact. We run separate regression models for each field to compare heterogeneity across fields. 
%We run a heterogeneous treatment effect analysis to investigate how the effects of jargon may vary across fields, so we run separate regression models for each field. 
Each unit of analysis is an abstract published in 2000-2014 labeled with only one or two subfields.

Two key independent variables are the fractions of discipline-specific words and senses in an abstract, with $c = 0.1$. For abstracts that have two subfields in the same analyzed parent field, we sum their type and sense jargon counts. Additional independent variables include time (three evenly-sized time bins within 2000-2014), length of abstract in tokens, number of authors, number of references in the article, number of subfields (one or two), and the venue's average citations per article. 

Citation count is an over-dispersed count variable, so we run a negative binomial regression to predict this outcome \cite{hilbe2011negative}. In some cases, jargon use has a significant positive relationship with citations, but the direction of this relationship differs across fields (Table~\ref{tab:success}, Appendix~\ref{appx:success}). 

Alternatively, interdisciplinary impact considers the subfield composition of articles citing a target abstract. We use \citet{leydesdorff2019inter}'s established formula, which they call $\text{DIV}$: 
\begin{equation}
\text{DIV}(\mathcal{C}) = \textcolor{violet}{\frac{n}{N}} \textcolor{olive}{(1-\text{Gini})} \textcolor{teal}{\sum_{i, j \in \mathcal{C}, i\neq j} \frac{d_{ij}}{n(n-1)}},
\end{equation}
where $\mathcal{C}$ is the set of subfields citing the abstract, $n = |\mathcal{C}|$, $N$ is the total number of subfields, and $d_{ij} = 1 - \cos(v_i, v_j)$, where $v$ are subfields vectorized using overall cross-subfield citation counts (Appendix~\ref{appx:success}). The \textcolor{violet}{first} component measures the fraction of citing subfields, the \textcolor{olive}{second} uses the Gini coefficient to calculate balance of citation counts among $\mathcal{C}$, and the \textcolor{teal}{third} incorporates subfield similarity \cite{leydesdorff2019inter,chen2022interdiscip,stirling1998economics}. We run ordinary least squares regression on abstracts that are cited by at least two subfields, with \textit{DIV} as the dependent variable. Discipline-specific words and senses have a negative relationship with \textit{DIV} across fields that have highly distinctive language norms (Table~\ref{tab:success}).

%Katie put down to 2 digits 
\begin{table}[t]
\centering
\resizebox{\columnwidth}{!}{%
\begin{tabular}{@{}lllrllr@{}}
\toprule
 & \multicolumn{3}{c}{\textbf{Citation count}} & \multicolumn{3}{c}{\textbf{Interdisciplinary impact (DIV)}} \\
\cmidrule(lr){2-4}\cmidrule(lr){5-7} 
\textbf{Field} & types & senses & \# obv. & types & senses & \# obv. \\ \midrule
Medicine & \ccell{cellgray}{-0.15***} & ~0.60*** & 1,137,923 & \ccell{cellgray}{-0.10***} & \ccell{cellgray}{-0.05***} & 589,641\\
Engineering & ~0.07 & ~0.64*** & 786,559 & \ccell{cellgray}{-0.09***} & \ccell{cellgray}{-0.15***} & 199,790 \\
Comp. sci. & \ccell{cellgray}{-0.87***} & ~0.71*** & 556,330 & \ccell{cellgray}{-0.12***} & \ccell{cellgray}{-0.11***} & 196,234\\
Biology & \ccell{cellgray}{-0.12***} & ~0.52*** & 824,768 & \ccell{cellgray}{-0.80***} & \ccell{cellgray}{-0.03***} & 481,103\\
Economics & ~0.15 & ~1.23*** & 454,215 & \ccell{cellgray}{-0.11***} & ~0.00 & 123,476\\
Physics & ~0.47*** & \ccell{cellgray}{-1.04***} & 648,729 & \ccell{cellgray}{-0.16***} & \ccell{cellgray}{-0.10***} & 203,009\\
Chemistry & \ccell{cellgray}{-1.36***} & \ccell{cellgray}{-2.32***} & 613,535 & \ccell{cellgray}{-0.10***} & \ccell{cellgray}{-0.08***} & 187,621\\
Mathematics & ~1.22*** & ~1.40*** & 363,369 & \ccell{cellgray}{-0.15***} & \ccell{cellgray}{-0.11***} & 128,482\\
Psychology & ~0.34*** & ~3.68*** & 261,102 & \ccell{cellgray}{-0.11***} & \ccell{cellgray}{-0.06***} & 133,319\\
Geology & \ccell{cellgray}{-0.42***} & ~0.83*** & 343,250 & \ccell{cellgray}{-0.13***} & \ccell{cellgray}{-0.13***} & 138,308\\
Sociology & ~1.18*** & ~2.24*** & 149,484 & \ccell{cellgray}{-0.08***} & ~0.01 & 56,088\\
Business & ~0.30** & ~2.71*** & 160,536 & \ccell{cellgray}{-0.11***} & \ccell{cellgray}{-0.04***} & 39,602\\
Environ. sci. & \ccell{cellgray}{-1.22***} & \ccell{cellgray}{-2.20***} & 137,862 & \ccell{cellgray}{-0.12***} & \ccell{cellgray}{-0.05***} & 49,199\\
Geography & ~0.17 & ~0.37 & 127,561 & \ccell{cellgray}{-0.10***} & \ccell{cellgray}{-0.04***} & 51,408\\
Material sci. & \ccell{cellgray}{-1.73***} & ~1.42*** & 149,602 & \ccell{cellgray}{-0.14***} & \ccell{cellgray}{-0.09***} & 45,445 \\
Philosophy & \ccell{cellgray}{-0.92***} & ~2.16*** & 68,512& \ccell{cellgray}{-0.03***} & ~0.06***& 10,559\\
Art & \ccell{cellgray}{-1.75***} & -2.30 & 68,220 & \ccell{cellgray}{-0.04***} & ~0.03 & 5,826\\
History & -0.27 & ~10.94*** & 47,910 & \ccell{cellgray}{-0.50***} & ~0.05 & 6,513\\
Political sci. & ~2.27*** & ~2.86*** & 44,994 & \ccell{cellgray}{-0.04**} & ~0.03 & 8,486\\
\bottomrule
\addlinespace[1ex]
\multicolumn{7}{l}{\textsuperscript{***}$p<0.001$, 
  \textsuperscript{**}$p<0.01$, 
  \textsuperscript{*}$p<0.05$ with Bonferroni correction.}
\end{tabular}%
}
\caption{The columns \textit{type} and \textit{sense} show regression coefficients for the fractions of discipline-specific words or senses in abstracts. The dependent variables are citation count and interdisciplinary impact (DIV). Significantly negative coefficients are highlighted, and \# obv. is the number of observations. Since dependent variables and their expected values differ across regressions, the magnitude of coefficients are not comparable.}
\label{tab:success}
\end{table}

Thus, though jargon has a varying relationship with citation counts, our regression results suggest that it may generally impede the forging of interdisciplinary connections.

\section{Related Work} 

Computational sociolinguistics often focuses on social media \cite{nguyen-etal-2016-survey}, with less attention on situation-dependent language varieties, or \textit{registers}, in scholarly communities \cite{agha2005registers}. Here, language differences can indicate different factions of authors and disciplinary approaches \cite{ngai2018discourse,west-portenoy-2016-delineating,sim-etal-2012-discovering}. In addition to our present work, a few studies have examined word meaning or use, such as semantic influence or novelty \cite{soni2021follow,soni2022predicting} and semantic uncertainty \cite{mcmahan2018ambiguity}. Research on lexical ambiguity in science also appears in education, with an emphasis on how to improve the teaching of overloaded terminology \cite{ryan1985language,cervetti2015factors}. Other NLP studies of science have predicted responses to articles \cite{yogatama-etal-2011-predicting}, measured impact and innovation \cite{gerow2018measuring, hofstra2020diversity,mckeown2016predicting}, and classified topics' rhetorical functions \cite{prabhakaran-etal-2016-predicting}. 

\section{Conclusion} 

We use data-driven, interpretable methods to identify jargon, defined as discipline-specific word types and senses, across science at scale. By identifying senses, we are able to recall more words labeled as associated with a field in Wiktionary than with word types alone. We then map language norms across subfields, showing that fields with distinctive word types differ from those with distinctive word senses. Finally, we analyze implications of jargon use for communication with out-groups. We find that supposedly general-purpose venues have varying expectations around abstracts' use of jargon depending on the field, and jargon is negatively related to interdisciplinary impact. This suggests a potential opportunity for the reconsideration of abstract writing norms, especially for venues that intend to bridge disciplines.

\section{Limitations}

Below, we outline several limitations of our work. 

\textbf{Data coverage}. Our claims are only valid for the datasets accessed in our study. We use the Microsoft Academic Graph \cite{sinha2015mag} and S2ORC, which is larger than other publicly-available scientific text corpora \cite{lo-etal-2020-s2orc}. However, these sources can differ from other collections of scientific text, because which journal/venues, sources, and resource types constitute ``science'' differs across academic literature search systems and databases \cite{gusenbauer2020search, ortega2014microsoft}. In particular, since a substantial portion of S2ORC comes from scrapes of arXiv and PubMed, its coverage of computer science and medicine is better than that of other fields \cite{lo-etal-2020-s2orc}. Also, our coverage is limited to English articles. Past work has shown that citation-based metrics of impact favor articles written in English, and articles from non-English-speaking countries have different citation patterns compared to others \cite{liang2013non,liu2018penalty,gonzalez2012impact}. Finally, we recognize that MAG field of study labels are contestable and imperfect. For example, less than two-thirds of \textit{ACL} articles are labeled as \textit{natural language processing}, and the most popular subfield in \textit{ICML} is \textit{mathematics} rather than \textit{machine learning}. 

\textbf{Token-level analyses}. Another limitation of our study is that many scholarly terms are not single words or tokens, but rather phrases. Phrases are somewhat accounted for by measuring words’ senses, since senses induced by language models reflect words’ in-context use, including their use in discipline-specific phrases. For example, Table~\ref{tab:top_sense_examples} shows that \textit{title} has a sense specific to stereochemistry, and in abstracts, this word often occurs in the phrases \textit{title reaction} or \textit{title compound}. Phrases containing distinctive words are also somewhat accounted for by measuring individual words in the phrase. However, phrase-level measurements of jargon would likely still be useful for improving interpretability and downstream applications of our metrics, and so discipline-specific phrases are a promising avenue for future work. 

\textbf{Compute.} Science of science is interdisciplinary and involves a range of organizations and institutions. Not all researchers will have easy access to the computuational resources needed to replicate our study or apply our approach to data of the same scale. The most resource intensive step of our pipeline is when ScholarBERT predicts each instance of a vocabulary word's top 5 substitutes across \textsc{Contemporary S2ORC} and \textsc{WikiSample}. This took approximately 90 GPU hours split across Nvidia RTX A6000 and Quadro RTX 8000 GPUs. ScholarBERT itself is a 770M-parameter BERT model \cite{scholarBERT2022}, and generally our compute infrastructure included machines with 64 to 128 cores and 512 to 1024 GB of RAM. 

\textbf{Social implications.} In \S\ref{sec:success}, we define ``success'' in two ways, both of which are based on citations. However, though citations are an important currency in science, they are imperfect signals of credit or impact. One article may cite another for reasons that span a range of significance, from brief mentions of related background to core motivation \cite{jurgens-etal-2018-measuring}. In addition, associations between jargon use and scientific success may differ as success is redefined using indicators beyond citations. For example, success could be defined beyond scientific communities, such as findings that lead to societal change, products, and use \cite{bornmann2013societal}. Finally, our study on the relationship between jargon and success is not causal, but associational and descriptive. 

\section{Ethical considerations}

\textbf{Data.} With regards to data privacy, the dataset we use, S2ORC, is not anonymized, since entries for each article includes a list of author names. Even with the removal of author names, data can easily be linked to authors since abstracts are published online with attribution. We don't use author information in our research, and our outputs are aggregated over subsets of data. Still, we acknowledge that science of science research involving author information has the risk of judging research productivity and quality using metrics that may deemphasize some forms of contribution and labor, systemically disadvantaging some demographic groups. In addition, we did not receive the explicit consent of authors to use their content for our study, though the harms of this are minimized since the type of science we study is inherently a public-facing endeavor. S2ORC is released under a CC BY-NC 4.0 license, and its intended use is for NLP task development and science of science analysis. Any derivatives we produce share the same intended use and license. 

\textbf{``Jargon''}. In this paper, we use \textit{jargon} to refer to sets of words that are specific to a discipline. \textit{Jargon} can be a neutral term when referring to scientific or technical language, but has negative connotations of being incomprehensible or undesirable when used to refer to community vernacular or entire language varieties. Thus, care should be taken when deciding when and how to use \textit{jargon} to refer to language.

\section{Acknowledgements}

We thank Misha Teplitskiy, Sandeep Soni, Isaac Bleaman, and Tal August for helpful conversations during the completion of this paper, and the Semantic Scholar team for their support in using and managing data. In addition, we thank our anonymous reviewers for their feedback. KK is grateful for the support of a Young Investigator Grant from the Allen Institute for Artificial Intelligence. 

\bibliography{anthology,custom}

\begin{thebibliography}{82}
\expandafter\ifx\csname natexlab\endcsname\relax\def\natexlab#1{#1}\fi

\bibitem[{Agha(2005)}]{agha2005registers}
Asif Agha. 2005.
\newblock \href {https://doi.org/https://doi.org/10.1002/9780470996522.ch2}
  {\emph{Registers of Language}}, chapter~2. John Wiley \& Sons, Ltd.

\bibitem[{Aharoni and Goldberg(2020)}]{aharoni-goldberg-2020-unsupervised}
Roee Aharoni and Yoav Goldberg. 2020.
\newblock \href {https://doi.org/10.18653/v1/2020.acl-main.692} {Unsupervised
  domain clusters in pretrained language models}.
\newblock In \emph{Proceedings of the 58th Annual Meeting of the Association
  for Computational Linguistics}, pages 7747--7763, Online. Association for
  Computational Linguistics.

\bibitem[{Androutsopoulos(2014)}]{androutsopoulos2014languaging}
Jannis Androutsopoulos. 2014.
\newblock \href {https://doi.org/https://doi.org/10.1016/j.dcm.2014.08.006}
  {Languaging when contexts collapse: Audience design in social networking}.
\newblock \emph{Discourse, Context \& Media}, 4-5:62--73.
\newblock Digital language practices in superdiversity.

\bibitem[{Attardi(2015)}]{Wikiextractor2015}
Giusepppe Attardi. 2015.
\newblock Wikiextractor.
\newblock \url{https://github.com/attardi/wikiextractor}.

\bibitem[{August et~al.(2020{\natexlab{a}})August, Card, Hsieh, Smith, and
  Reinecke}]{august2020explain}
Tal August, Dallas Card, Gary Hsieh, Noah~A. Smith, and Katharina Reinecke.
  2020{\natexlab{a}}.
\newblock \href {https://doi.org/10.1145/3313831.3376524} {Explain like {I} am
  a scientist: The linguistic barriers of entry to r/science}.
\newblock In \emph{Proceedings of the 2020 CHI Conference on Human Factors in
  Computing Systems}, CHI '20, page 1–12, New York, NY, USA. Association for
  Computing Machinery.

\bibitem[{August et~al.(2020{\natexlab{b}})August, Kim, Reinecke, and
  Smith}]{august-etal-2020-writing}
Tal August, Lauren Kim, Katharina Reinecke, and Noah~A. Smith.
  2020{\natexlab{b}}.
\newblock \href {https://doi.org/10.18653/v1/2020.emnlp-main.429} {Writing
  strategies for science communication: Data and computational analysis}.
\newblock In \emph{Proceedings of the 2020 Conference on Empirical Methods in
  Natural Language Processing (EMNLP)}, pages 5327--5344, Online. Association
  for Computational Linguistics.

\bibitem[{August et~al.(2022{\natexlab{a}})August, Reinecke, and
  Smith}]{august-etal-2022-generating}
Tal August, Katharina Reinecke, and Noah~A. Smith. 2022{\natexlab{a}}.
\newblock \href {https://doi.org/10.18653/v1/2022.acl-long.569} {Generating
  scientific definitions with controllable complexity}.
\newblock In \emph{Proceedings of the 60th Annual Meeting of the Association
  for Computational Linguistics (Volume 1: Long Papers)}, pages 8298--8317,
  Dublin, Ireland. Association for Computational Linguistics.

\bibitem[{August et~al.(2022{\natexlab{b}})August, Wang, Bragg, Hearst, Head,
  and Lo}]{august2022paperplain}
Tal August, Lucy~Lu Wang, Jonathan Bragg, Marti~A. Hearst, Andrew Head, and
  Kyle Lo. 2022{\natexlab{b}}.
\newblock \href {https://doi.org/10.48550/ARXIV.2203.00130} {Paper plain:
  Making medical research papers approachable to healthcare consumers with
  natural language processing}.

\bibitem[{Bell(1984)}]{bell_1984}
Allan Bell. 1984.
\newblock \href {https://doi.org/10.1017/S004740450001037X} {Language style as
  audience design}.
\newblock \emph{Language in Society}, 13(2):145–204.

\bibitem[{Bell(2002)}]{bell_2002}
Allan Bell. 2002.
\newblock \href {https://doi.org/10.1017/CBO9780511613258.010} {\emph{Back in
  style: reworking audience design}}, page 139–169. Cambridge University
  Press.

\bibitem[{Beltagy et~al.(2019)Beltagy, Lo, and
  Cohan}]{beltagy-etal-2019-scibert}
Iz~Beltagy, Kyle Lo, and Arman Cohan. 2019.
\newblock \href {https://doi.org/10.18653/v1/D19-1371} {{S}ci{BERT}: A
  pretrained language model for scientific text}.
\newblock In \emph{Proceedings of the 2019 Conference on Empirical Methods in
  Natural Language Processing and the 9th International Joint Conference on
  Natural Language Processing (EMNLP-IJCNLP)}, pages 3615--3620, Hong Kong,
  China. Association for Computational Linguistics.

\bibitem[{Blondel et~al.(2008)Blondel, Guillaume, Lambiotte, and
  Lefebvre}]{Blondel_2008}
Vincent~D Blondel, Jean-Loup Guillaume, Renaud Lambiotte, and Etienne Lefebvre.
  2008.
\newblock \href {https://doi.org/10.1088/1742-5468/2008/10/P10008} {Fast
  unfolding of communities in large networks}.
\newblock \emph{Journal of Statistical Mechanics: Theory and Experiment},
  2008(10):P10008.

\bibitem[{Bornmann(2013)}]{bornmann2013societal}
Lutz Bornmann. 2013.
\newblock \href {https://doi.org/https://doi.org/10.1002/asi.22803} {What is
  societal impact of research and how can it be assessed? {A} literature
  survey}.
\newblock \emph{Journal of the American Society for Information Science and
  Technology}, 64(2):217--233.

\bibitem[{Boyack et~al.(2005)Boyack, Klavans, and
  B{\"o}rner}]{boyack2005mapping}
Kevin~W Boyack, Richard Klavans, and Katy B{\"o}rner. 2005.
\newblock Mapping the backbone of science.
\newblock \emph{Scientometrics}, 64(3):351--374.

\bibitem[{Breusch and Pagan(1979)}]{breusch1979test}
T.~S. Breusch and A.~R. Pagan. 1979.
\newblock \href {http://www.jstor.org/stable/1911963} {A simple test for
  heteroscedasticity and random coefficient variation}.
\newblock \emph{Econometrica}, 47(5):1287--1294.

\bibitem[{Bullinaria and Levy(2007)}]{bullinaria2007extracting}
John~A Bullinaria and Joseph~P Levy. 2007.
\newblock Extracting semantic representations from word co-occurrence
  statistics: A computational study.
\newblock \emph{Behavior Research Methods}, 39(3):510--526.

\bibitem[{Cameron and Trivedi(2013)}]{cameron2013regression}
A~Colin Cameron and Pravin~K Trivedi. 2013.
\newblock \emph{Regression Analysis of Count Data}.
\newblock Cambridge University Press.

\bibitem[{Cervetti et~al.(2015)Cervetti, Hiebert, Pearson, and
  McClung}]{cervetti2015factors}
Gina~N. Cervetti, Elfrieda~H. Hiebert, P.~David Pearson, and Nicola~A. McClung.
  2015.
\newblock \href {https://doi.org/10.1177/1086296X15615363} {Factors that
  influence the difficulty of science words}.
\newblock \emph{Journal of Literacy Research}, 47(2):153--185.

\bibitem[{Chen et~al.(2022)Chen, Song, Shu, and
  Larivi\`{e}re}]{chen2022interdiscip}
Shiji Chen, Yanhui Song, Fei Shu, and Vincent Larivi\`{e}re. 2022.
\newblock \href {https://doi.org/10.1007/s11192-022-04338-1}
  {Interdisciplinarity and impact: The effects of the citation time window}.
\newblock \emph{Scientometrics}, 127(5):2621–2642.

\bibitem[{Dagan et~al.(1993)Dagan, Marcus, and
  Markovitch}]{dagan-etal-1993-contextual}
Ido Dagan, Shaul Marcus, and Shaul Markovitch. 1993.
\newblock \href {https://doi.org/10.3115/981574.981596} {Contextual word
  similarity and estimation from sparse data}.
\newblock In \emph{31st Annual Meeting of the Association for Computational
  Linguistics}, pages 164--171, Columbus, Ohio, USA. Association for
  Computational Linguistics.

\bibitem[{de~Carli and Pereira(2017)}]{de2017multidisciplinarity}
Gabriel~Jos{\'e} de~Carli and Tiago~Campos Pereira. 2017.
\newblock Multidisciplinarity: Widen discipline span of nature papers.
\newblock \emph{Nature}, 545(7654):289--289.

\bibitem[{Eyal et~al.(2022)Eyal, Sadde, Taub-Tabib, and
  Goldberg}]{eyal-etal-2022-large}
Matan Eyal, Shoval Sadde, Hillel Taub-Tabib, and Yoav Goldberg. 2022.
\newblock \href {https://doi.org/10.18653/v1/2022.acl-long.325} {Large scale
  substitution-based word sense induction}.
\newblock In \emph{Proceedings of the 60th Annual Meeting of the Association
  for Computational Linguistics (Volume 1: Long Papers)}, pages 4738--4752,
  Dublin, Ireland. Association for Computational Linguistics.

\bibitem[{Fortunato et~al.(2018)Fortunato, Bergstrom, Börner, Evans, Helbing,
  Milojević, Petersen, Radicchi, Sinatra, Uzzi, Vespignani, Waltman, Wang, and
  Barabási}]{fortunato2018science}
Santo Fortunato, Carl~T. Bergstrom, Katy Börner, James~A. Evans, Dirk Helbing,
  Staša Milojević, Alexander~M. Petersen, Filippo Radicchi, Roberta Sinatra,
  Brian Uzzi, Alessandro Vespignani, Ludo Waltman, Dashun Wang, and
  Albert-László Barabási. 2018.
\newblock \href {https://doi.org/10.1126/science.aao0185} {Science of science}.
\newblock \emph{Science}, 359(6379):eaao0185.

\bibitem[{Foster et~al.(2015)Foster, Rzhetsky, and Evans}]{foster2015tradition}
Jacob~G. Foster, Andrey Rzhetsky, and James~A. Evans. 2015.
\newblock \href {https://doi.org/10.1177/0003122415601618} {Tradition and
  innovation in scientists’ research strategies}.
\newblock \emph{American Sociological Review}, 80(5):875--908.

\bibitem[{Freeling et~al.(2019)Freeling, Doubleday, and
  Connell}]{freeling2019how}
Benjamin Freeling, Zoë~A. Doubleday, and Sean~D. Connell. 2019.
\newblock \href {https://doi.org/10.1073/pnas.1819937116} {How can we boost the
  impact of publications? {T}ry better writing}.
\newblock \emph{Proceedings of the National Academy of Sciences},
  116(2):341--343.

\bibitem[{Freeling et~al.(2021)Freeling, Doubleday, Dry, Semmler, and
  Connell}]{freeling2021better}
Benjamin~S. Freeling, Zoë~A. Doubleday, Matthew~J. Dry, Carolyn Semmler, and
  Sean~D. Connell. 2021.
\newblock \href {https://doi.org/10.3389/fpsyg.2021.714321} {Better writing in
  scientific publications builds reader confidence and understanding}.
\newblock \emph{Frontiers in Psychology}, 12.

\bibitem[{Gardner et~al.(2021)Gardner, Merrill, Dodge, Peters, Ross, Singh, and
  Smith}]{gardner-etal-2021-competency}
Matt Gardner, William Merrill, Jesse Dodge, Matthew Peters, Alexis Ross, Sameer
  Singh, and Noah~A. Smith. 2021.
\newblock \href {https://doi.org/10.18653/v1/2021.emnlp-main.135} {Competency
  problems: On finding and removing artifacts in language data}.
\newblock In \emph{Proceedings of the 2021 Conference on Empirical Methods in
  Natural Language Processing}, pages 1801--1813, Online and Punta Cana,
  Dominican Republic. Association for Computational Linguistics.

\bibitem[{Gerow et~al.(2018)Gerow, Hu, Boyd-Graber, Blei, and
  Evans}]{gerow2018measuring}
Aaron Gerow, Yuening Hu, Jordan Boyd-Graber, David~M. Blei, and James~A. Evans.
  2018.
\newblock \href {https://doi.org/10.1073/pnas.1719792115} {Measuring discursive
  influence across scholarship}.
\newblock \emph{Proceedings of the National Academy of Sciences},
  115(13):3308--3313.

\bibitem[{Gonz{\'a}lez-Alcaide et~al.(2012)Gonz{\'a}lez-Alcaide,
  Valderrama-Zuri{\'a}n, and Aleixandre-Benavent}]{gonzalez2012impact}
Gregorio Gonz{\'a}lez-Alcaide, Juan~Carlos Valderrama-Zuri{\'a}n, and Rafael
  Aleixandre-Benavent. 2012.
\newblock The impact factor in non-{E}nglish-speaking countries.
\newblock \emph{Scientometrics}, 92(2):297--311.

\bibitem[{Gusenbauer and Haddaway(2020)}]{gusenbauer2020search}
Michael Gusenbauer and Neal~R. Haddaway. 2020.
\newblock \href {https://doi.org/https://doi.org/10.1002/jrsm.1378} {Which
  academic search systems are suitable for systematic reviews or meta-analyses?
  {E}valuating retrieval qualities of {G}oogle {S}cholar, {PubMed}, and 26
  other resources}.
\newblock \emph{Research Synthesis Methods}, 11(2):181--217.

\bibitem[{Head et~al.(2021)Head, Lo, Kang, Fok, Skjonsberg, Weld, and
  Hearst}]{head2021augmenting}
Andrew Head, Kyle Lo, Dongyeop Kang, Raymond Fok, Sam Skjonsberg, Daniel~S.
  Weld, and Marti~A. Hearst. 2021.
\newblock \href {https://doi.org/10.1145/3411764.3445648} {Augmenting
  scientific papers with just-in-time, position-sensitive definitions of terms
  and symbols}.
\newblock In \emph{Proceedings of the 2021 CHI Conference on Human Factors in
  Computing Systems}, CHI '21, New York, NY, USA. Association for Computing
  Machinery.

\bibitem[{Hilbe(2011)}]{hilbe2011negative}
Joseph~M Hilbe. 2011.
\newblock \emph{Negative binomial regression}.
\newblock Cambridge University Press.

\bibitem[{Hofstra et~al.(2020)Hofstra, Kulkarni, Galvez, He, Jurafsky, and
  McFarland}]{hofstra2020diversity}
Bas Hofstra, Vivek~V. Kulkarni, Sebastian Munoz-Najar Galvez, Bryan He, Dan
  Jurafsky, and Daniel~A. McFarland. 2020.
\newblock \href {https://doi.org/10.1073/pnas.1915378117} {The
  diversity-innovation paradox in science}.
\newblock \emph{Proceedings of the National Academy of Sciences},
  117(17):9284--9291.

\bibitem[{Hong et~al.(2022)Hong, Ajith, Pauloski, Duede, Malamud, Magoulas,
  Chard, and Foster}]{scholarBERT2022}
Zhi Hong, Aswathy Ajith, Gregory Pauloski, Eamon Duede, Carl Malamud, Roger
  Magoulas, Kyle Chard, and Ian Foster. 2022.
\newblock \href {https://doi.org/10.48550/ARXIV.2205.11342} {Scholar{BERT}:
  Bigger is not always better}.

\bibitem[{Jurgens et~al.(2018)Jurgens, Kumar, Hoover, McFarland, and
  Jurafsky}]{jurgens-etal-2018-measuring}
David Jurgens, Srijan Kumar, Raine Hoover, Dan McFarland, and Dan Jurafsky.
  2018.
\newblock \href {https://doi.org/10.1162/tacl_a_00028} {Measuring the evolution
  of a scientific field through citation frames}.
\newblock \emph{Transactions of the Association for Computational Linguistics},
  6:391--406.

\bibitem[{Kim et~al.(2016)Kim, Hullman, Burgess, and
  Adar}]{kim-etal-2016-simplescience}
Yea-Seul Kim, Jessica Hullman, Matthew Burgess, and Eytan Adar. 2016.
\newblock \href {https://doi.org/10.18653/v1/D16-1114} {{S}imple{S}cience:
  Lexical simplification of scientific terminology}.
\newblock In \emph{Proceedings of the 2016 Conference on Empirical Methods in
  Natural Language Processing}, pages 1066--1071, Austin, Texas. Association
  for Computational Linguistics.

\bibitem[{Koopman(2011)}]{koopman2011scientific}
Ann Koopman. 2011.
\newblock Nature launches new open access journal: Scientific reports.
\newblock \emph{Thomas Jefferson University Library News}.

\bibitem[{Labov(1973)}]{labov1973sociolinguistic}
William Labov. 1973.
\newblock \emph{Sociolinguistic patterns}.
\newblock 4. University of Pennsylvania Press.

\bibitem[{Larivière and Gingras(2010)}]{lariviere2010relationship}
Vincent Larivière and Yves Gingras. 2010.
\newblock \href {https://doi.org/https://doi.org/10.1002/asi.21226} {On the
  relationship between interdisciplinarity and scientific impact}.
\newblock \emph{Journal of the American Society for Information Science and
  Technology}, 61(1):126--131.

\bibitem[{Levy et~al.(2015)Levy, Goldberg, and
  Dagan}]{levy-etal-2015-improving}
Omer Levy, Yoav Goldberg, and Ido Dagan. 2015.
\newblock \href {https://doi.org/10.1162/tacl_a_00134} {Improving
  distributional similarity with lessons learned from word embeddings}.
\newblock \emph{Transactions of the Association for Computational Linguistics},
  3:211--225.

\bibitem[{Leydesdorff et~al.(2019)Leydesdorff, Wagner, and
  Bornmann}]{leydesdorff2019inter}
Loet Leydesdorff, Caroline~S. Wagner, and Lutz Bornmann. 2019.
\newblock \href {https://doi.org/https://doi.org/10.1016/j.joi.2018.12.006}
  {Interdisciplinarity as diversity in citation patterns among journals:
  Rao-stirling diversity, relative variety, and the gini coefficient}.
\newblock \emph{Journal of Informetrics}, 13(1):255--269.

\bibitem[{Liang et~al.(2013)Liang, Rousseau, and Zhong}]{liang2013non}
Liming Liang, Ronald Rousseau, and Zhen Zhong. 2013.
\newblock Non-{E}nglish journals and papers in physics and chemistry: Bias in
  citations?
\newblock \emph{Scientometrics}, 95(1):333--350.

\bibitem[{Liu et~al.(2018)Liu, Hu, Tang, and Liu}]{liu2018penalty}
Fang Liu, Guangyuan Hu, Li~Tang, and Weishu Liu. 2018.
\newblock The penalty of containing more non-{E}nglish articles.
\newblock \emph{Scientometrics}, 114(1):359--366.

\bibitem[{Liu et~al.(2022)Liu, Medlar, and G{\l}owacka}]{liu2022lexical}
Yang Liu, Alan Medlar, and Dorota G{\l}owacka. 2022.
\newblock Lexical ambiguity detection in professional discourse.
\newblock \emph{Information Processing \& Management}, 59(5):103000.

\bibitem[{Lo et~al.(2020)Lo, Wang, Neumann, Kinney, and
  Weld}]{lo-etal-2020-s2orc}
Kyle Lo, Lucy~Lu Wang, Mark Neumann, Rodney Kinney, and Daniel Weld. 2020.
\newblock \href {https://doi.org/10.18653/v1/2020.acl-main.447} {{S}2{ORC}: The
  semantic scholar open research corpus}.
\newblock In \emph{Proceedings of the 58th Annual Meeting of the Association
  for Computational Linguistics}, pages 4969--4983, Online. Association for
  Computational Linguistics.

\bibitem[{Lucy and Bamman(2021)}]{lucy-bamman-2021-characterizing}
Li~Lucy and David Bamman. 2021.
\newblock \href {https://doi.org/10.1162/tacl_a_00383} {Characterizing
  {E}nglish variation across social media communities with {BERT}}.
\newblock \emph{Transactions of the Association for Computational Linguistics},
  9:538--556.

\bibitem[{Lui and Baldwin(2012)}]{lui-baldwin-2012-langid}
Marco Lui and Timothy Baldwin. 2012.
\newblock \href {https://aclanthology.org/P12-3005} {langid.py: An
  off-the-shelf language identification tool}.
\newblock In \emph{Proceedings of the {ACL} 2012 System Demonstrations}, pages
  25--30, Jeju Island, Korea. Association for Computational Linguistics.

\bibitem[{Mart{\'\i}nez and Mammola(2021)}]{martinez2021specialized}
Alejandro Mart{\'\i}nez and Stefano Mammola. 2021.
\newblock Specialized terminology reduces the number of citations of scientific
  papers.
\newblock \emph{Proceedings of the Royal Society B}, 288(1948):20202581.

\bibitem[{McKeown et~al.(2016)McKeown, Daume~III, Chaturvedi, Paparrizos,
  Thadani, Barrio, Biran, Bothe, Collins, Fleischmann, Gravano, Jha, King,
  McInerney, Moon, Neelakantan, O'Seaghdha, Radev, Templeton, and
  Teufel}]{mckeown2016predicting}
Kathy McKeown, Hal Daume~III, Snigdha Chaturvedi, John Paparrizos, Kapil
  Thadani, Pablo Barrio, Or~Biran, Suvarna Bothe, Michael Collins, Kenneth~R.
  Fleischmann, Luis Gravano, Rahul Jha, Ben King, Kevin McInerney, Taesun Moon,
  Arvind Neelakantan, Diarmuid O'Seaghdha, Dragomir Radev, Clay Templeton, and
  Simone Teufel. 2016.
\newblock \href {https://doi.org/https://doi.org/10.1002/asi.23612} {Predicting
  the impact of scientific concepts using full-text features}.
\newblock \emph{Journal of the Association for Information Science and
  Technology}, 67(11):2684--2696.

\bibitem[{McMahan and Evans(2018)}]{mcmahan2018ambiguity}
Peter McMahan and James Evans. 2018.
\newblock \href {https://doi.org/10.1086/701298} {Ambiguity and engagement}.
\newblock \emph{American Journal of Sociology}, 124(3):860--912.

\bibitem[{Mesgari et~al.(2015)Mesgari, Okoli, Mehdi, Nielsen, and
  Lanam{\"a}ki}]{mesgari2015sum}
Mostafa Mesgari, Chitu Okoli, Mohamad Mehdi, Finn~{\AA}rup Nielsen, and Arto
  Lanam{\"a}ki. 2015.
\newblock “the sum of all human knowledge”: A systematic review of
  scholarly research on the content of w ikipedia.
\newblock \emph{Journal of the Association for Information Science and
  Technology}, 66(2):219--245.

\bibitem[{Murthy et~al.(2022)Murthy, Lo, King, Bhagavatula, Kuehl, Johnson,
  Borchardt, Weld, Hope, and Downey}]{murthy2022accord}
Sonia~K Murthy, Kyle Lo, Daniel King, Chandra Bhagavatula, Bailey Kuehl, Sophie
  Johnson, Jonathan Borchardt, Daniel~S Weld, Tom Hope, and Doug Downey. 2022.
\newblock Accord: A multi-document approach to generating diverse descriptions
  of scientific concepts.
\newblock In \emph{Proceedings of the EMNLP 2022 System Demonstrations}.

\bibitem[{Ndubuisi-Obi et~al.(2019)Ndubuisi-Obi, Ghosh, and
  Jurgens}]{ndubuisi-obi-etal-2019-wetin}
Innocent Ndubuisi-Obi, Sayan Ghosh, and David Jurgens. 2019.
\newblock \href {https://doi.org/10.18653/v1/P19-1625} {Wetin dey with these
  comments? modeling sociolinguistic factors affecting code-switching behavior
  in nigerian online discussions}.
\newblock In \emph{Proceedings of the 57th Annual Meeting of the Association
  for Computational Linguistics}, pages 6204--6214, Florence, Italy.
  Association for Computational Linguistics.

\bibitem[{Newman(2016)}]{newman2016equivalence}
Mark~EJ Newman. 2016.
\newblock Equivalence between modularity optimization and maximum likelihood
  methods for community detection.
\newblock \emph{Physical Review E}, 94(5):052315.

\bibitem[{Ngai et~al.(2018)Ngai, Singh, and Koon}]{ngai2018discourse}
Sing Bik~Cindy Ngai, Rita~Gill Singh, and Alex~Chun Koon. 2018.
\newblock \href {https://doi.org/10.1371/journal.pone.0205417} {A discourse
  analysis of the macro-structure, metadiscoursal and microdiscoursal features
  in the abstracts of research articles across multiple science disciplines}.
\newblock \emph{PLOS ONE}, 13(10):1--21.

\bibitem[{Nguyen et~al.(2016)Nguyen, Do{\u{g}}ru{\"o}z, Ros{\'e}, and
  de~Jong}]{nguyen-etal-2016-survey}
Dong Nguyen, A.~Seza Do{\u{g}}ru{\"o}z, Carolyn~P. Ros{\'e}, and Franciska
  de~Jong. 2016.
\newblock \href {https://doi.org/10.1162/COLI_a_00258} {{S}urvey: Computational
  sociolinguistics: A {S}urvey}.
\newblock \emph{Computational Linguistics}, 42(3):537--593.

\bibitem[{Okamura(2019)}]{okamura2019interdisciplinarity}
Keisuke Okamura. 2019.
\newblock Interdisciplinarity revisited: evidence for research impact and
  dynamism.
\newblock \emph{Palgrave Communications}, 5(1):1--9.

\bibitem[{Ortega and Aguillo(2014)}]{ortega2014microsoft}
José~Luis Ortega and Isidro~F. Aguillo. 2014.
\newblock \href {https://doi.org/https://doi.org/10.1002/asi.23036} {Microsoft
  {a}cademic search and google {s}cholar citations: Comparative analysis of
  author profiles}.
\newblock \emph{Journal of the Association for Information Science and
  Technology}, 65(6):1149--1156.

\bibitem[{Pavalanathan and Eisenstein(2015)}]{pavalanathan2015audience}
Umashanthi Pavalanathan and Jacob Eisenstein. 2015.
\newblock \href {https://doi.org/10.1215/00031283-3130324} {{Audience-Modulated
  Variation in Online Social Media}}.
\newblock \emph{American Speech}, 90(2):187--213.

\bibitem[{Peng et~al.(2021)Peng, Ke, Budak, Romero, and Ahn}]{peng2021neural}
Hao Peng, Qing Ke, Ceren Budak, Daniel~M Romero, and Yong-Yeol Ahn. 2021.
\newblock Neural embeddings of scholarly periodicals reveal complex
  disciplinary organizations.
\newblock \emph{Science Advances}, 7(17):eabb9004.

\bibitem[{Plavén-Sigray et~al.(2017)Plavén-Sigray, Matheson, Schiffler, and
  Thompson}]{plaven_sigray2017research}
Pontus Plavén-Sigray, Granville~James Matheson, Björn~Christian Schiffler,
  and William~Hedley Thompson. 2017.
\newblock \href {https://doi.org/10.7554/eLife.27725} {Research: The
  readability of scientific texts is decreasing over time}.
\newblock \emph{eLife}, 6:e27725.

\bibitem[{Prabhakaran et~al.(2016)Prabhakaran, Hamilton, McFarland, and
  Jurafsky}]{prabhakaran-etal-2016-predicting}
Vinodkumar Prabhakaran, William~L. Hamilton, Dan McFarland, and Dan Jurafsky.
  2016.
\newblock \href {https://doi.org/10.18653/v1/P16-1111} {Predicting the rise and
  fall of scientific topics from trends in their rhetorical framing}.
\newblock In \emph{Proceedings of the 54th Annual Meeting of the Association
  for Computational Linguistics (Volume 1: Long Papers)}, pages 1170--1180,
  Berlin, Germany. Association for Computational Linguistics.

\bibitem[{Rakedzon et~al.(2017)Rakedzon, Segev, Chapnik, Yosef, and
  Baram-Tsabari}]{rakedzon2017automatic}
Tzipora Rakedzon, Elad Segev, Noam Chapnik, Roy Yosef, and Ayelet
  Baram-Tsabari. 2017.
\newblock \href {https://doi.org/10.1371/journal.pone.0181742} {Automatic
  jargon identifier for scientists engaging with the public and science
  communication educators}.
\newblock \emph{PLOS ONE}, 12(8):1--13.

\bibitem[{Ramesh~Kashyap et~al.(2021)Ramesh~Kashyap, Hazarika, Kan, and
  Zimmermann}]{ramesh-kashyap-etal-2021-domain}
Abhinav Ramesh~Kashyap, Devamanyu Hazarika, Min-Yen Kan, and Roger Zimmermann.
  2021.
\newblock \href {https://doi.org/10.18653/v1/2021.naacl-main.147} {Domain
  divergences: A survey and empirical analysis}.
\newblock In \emph{Proceedings of the 2021 Conference of the North American
  Chapter of the Association for Computational Linguistics: Human Language
  Technologies}, pages 1830--1849, Online. Association for Computational
  Linguistics.

\bibitem[{Reagle and Koerner(2020)}]{reagle2020wikipedia}
Joseph Reagle and Jackie Koerner. 2020.
\newblock \emph{Wikipedia@ 20: Stories of an incomplete revolution}.
\newblock The MIT Press.

\bibitem[{Rosvall and Bergstrom(2008)}]{rosvall2008maps}
Martin Rosvall and Carl~T Bergstrom. 2008.
\newblock Maps of random walks on complex networks reveal community structure.
\newblock \emph{Proceedings of the national academy of sciences},
  105(4):1118--1123.

\bibitem[{Ryan(1985)}]{ryan1985language}
Janet~N. Ryan. 1985.
\newblock \href {https://doi.org/10.1021/ed062p1098} {The language gap: Common
  words with technical meanings}.
\newblock \emph{Journal of Chemical Education}, 62(12):1098.

\bibitem[{{Satishkumar} et~al.(2000){Satishkumar}, {Thomas}, {Govindaraj}, and
  {Rao}}]{Satishkumar2000applied}
B.~C. {Satishkumar}, P.~John {Thomas}, A.~{Govindaraj}, and C.~N.~R. {Rao}.
  2000.
\newblock \href {https://doi.org/10.1063/1.1319185} {{Y-junction carbon
  nanotubes}}.
\newblock \emph{Applied Physics Letters}, 77(16):2530.

\bibitem[{Sim et~al.(2012)Sim, Smith, and Smith}]{sim-etal-2012-discovering}
Yanchuan Sim, Noah~A. Smith, and David~A. Smith. 2012.
\newblock \href {https://aclanthology.org/W12-3203} {Discovering factions in
  the computational linguistics community}.
\newblock In \emph{Proceedings of the {ACL}-2012 Special Workshop on
  Rediscovering 50 Years of Discoveries}, pages 22--32, Jeju Island, Korea.
  Association for Computational Linguistics.

\bibitem[{Sinha et~al.(2015)Sinha, Shen, Song, Ma, Eide, Hsu, and
  Wang}]{sinha2015mag}
Arnab Sinha, Zhihong Shen, Yang Song, Hao Ma, Darrin Eide, Bo-June~(Paul) Hsu,
  and Kuansan Wang. 2015.
\newblock \href {https://doi.org/10.1145/2740908.2742839} {An overview of
  microsoft academic service (mas) and applications}.
\newblock In \emph{Proceedings of the 24th International Conference on World
  Wide Web}, WWW '15 Companion, page 243–246, New York, NY, USA. Association
  for Computing Machinery.

\bibitem[{Soni et~al.(2022)Soni, Bamman, and Eisenstein}]{soni2022predicting}
Sandeep Soni, David Bamman, and Jacob Eisenstein. 2022.
\newblock \href {https://doi.org/10.48550/ARXIV.2210.13628} {Predicting
  long-term citations from short-term linguistic influence}.

\bibitem[{Soni et~al.(2021)Soni, Lerman, and Eisenstein}]{soni2021follow}
Sandeep Soni, Kristina Lerman, and Jacob Eisenstein. 2021.
\newblock \href {https://doi.org/https://doi.org/10.1002/asi.24421} {Follow the
  leader: Documents on the leading edge of semantic change get more citations}.
\newblock \emph{Journal of the Association for Information Science and
  Technology}, 72(4):478--492.

\bibitem[{Stirling(1998)}]{stirling1998economics}
Andrew Stirling. 1998.
\newblock On the economics and analysis of diversity.
\newblock \emph{Science Policy Research Unit (SPRU), Electronic Working Papers
  Series, Paper}, 28:1--156.

\bibitem[{Vadapalli et~al.(2018)Vadapalli, Syed, Prabhu, Srinivasan, and
  Varma}]{vadapalli-etal-2018-science}
Raghuram Vadapalli, Bakhtiyar Syed, Nishant Prabhu, Balaji~Vasan Srinivasan,
  and Vasudeva Varma. 2018.
\newblock \href {https://doi.org/10.18653/v1/D18-2028} {When science journalism
  meets artificial intelligence : An interactive demonstration}.
\newblock In \emph{Proceedings of the 2018 Conference on Empirical Methods in
  Natural Language Processing: System Demonstrations}, pages 163--168,
  Brussels, Belgium. Association for Computational Linguistics.

\bibitem[{Van~Noorden(2015)}]{van2015interdisciplinary}
Richard Van~Noorden. 2015.
\newblock Interdisciplinary research by the numbers.
\newblock \emph{Nature}, 525(7569):306--307.

\bibitem[{Varmus et~al.(2000)Varmus, Brown, and Eisen}]{varmus2000open}
Harold Varmus, Patrick Brown, and Michael Eisen. 2000.
\newblock \href {https://plos.org/open-letter/} {Open letter}.
\newblock \emph{PLOS One}.

\bibitem[{Vilhena et~al.(2014)Vilhena, Foster, Rosvall, West, Evans, and
  Bergstrom}]{vilhena2014finding}
Daril~A Vilhena, Jacob~G Foster, Martin Rosvall, Jevin~D West, James Evans, and
  Carl~T Bergstrom. 2014.
\newblock Finding cultural holes: How structure and culture diverge in networks
  of scholarly communication.
\newblock \emph{Sociological Science}, 1:221.

\bibitem[{Wang and Barabási(2021)}]{wang_barabasi_2021}
Dashun Wang and Albert-László Barabási. 2021.
\newblock \href {https://doi.org/10.1017/9781108610834.004} {\emph{The
  h-Index}}, page 17–27. Cambridge University Press.

\bibitem[{Wang et~al.(2019)Wang, Shen, Huang, Wu, Eide, Dong, Qian, Kanakia,
  Chen, and Rogahn}]{wang2019mag}
Kuansan Wang, Zhihong Shen, Chiyuan Huang, Chieh-Han Wu, Darrin Eide, Yuxiao
  Dong, Junjie Qian, Anshul Kanakia, Alvin Chen, and Richard Rogahn. 2019.
\newblock \href {https://doi.org/10.3389/fdata.2019.00045} {A review of
  {M}icrosoft {A}cademic {S}ervices for science of science studies}.
\newblock \emph{Frontiers in Big Data}, 2.

\bibitem[{West and Portenoy(2016)}]{west-portenoy-2016-delineating}
Jevin West and Jason Portenoy. 2016.
\newblock \href {https://aclanthology.org/W16-1508} {Delineating fields using
  mathematical jargon}.
\newblock In \emph{Proceedings of the Joint Workshop on Bibliometric-enhanced
  Information Retrieval and Natural Language Processing for Digital Libraries
  ({BIRNDL})}, pages 63--71.

\bibitem[{Yogatama et~al.(2011)Yogatama, Heilman, O{'}Connor, Dyer, Routledge,
  and Smith}]{yogatama-etal-2011-predicting}
Dani Yogatama, Michael Heilman, Brendan O{'}Connor, Chris Dyer, Bryan~R.
  Routledge, and Noah~A. Smith. 2011.
\newblock \href {https://aclanthology.org/D11-1055} {Predicting a scientific
  community{'}s response to an article}.
\newblock In \emph{Proceedings of the 2011 Conference on Empirical Methods in
  Natural Language Processing}, pages 594--604, Edinburgh, Scotland, UK.
  Association for Computational Linguistics.

\bibitem[{Zhang et~al.(2017)Zhang, Hamilton, Danescu-Niculescu-Mizil, Jurafsky,
  and Leskovec}]{Zhang2017community}
Justine Zhang, William Hamilton, Cristian Danescu-Niculescu-Mizil, Dan
  Jurafsky, and Jure Leskovec. 2017.
\newblock \href {https://ojs.aaai.org/index.php/ICWSM/article/view/14904}
  {Community identity and user engagement in a multi-community landscape}.
\newblock \emph{Proceedings of the International AAAI Conference on Web and
  Social Media}, 11(1):377--386.

\end{thebibliography}
\bibliographystyle{acl_natbib}

\appendix

\section{Fields of study}\label{appx:fos}

\begin{figure}[t]
    \includegraphics[width=\columnwidth]{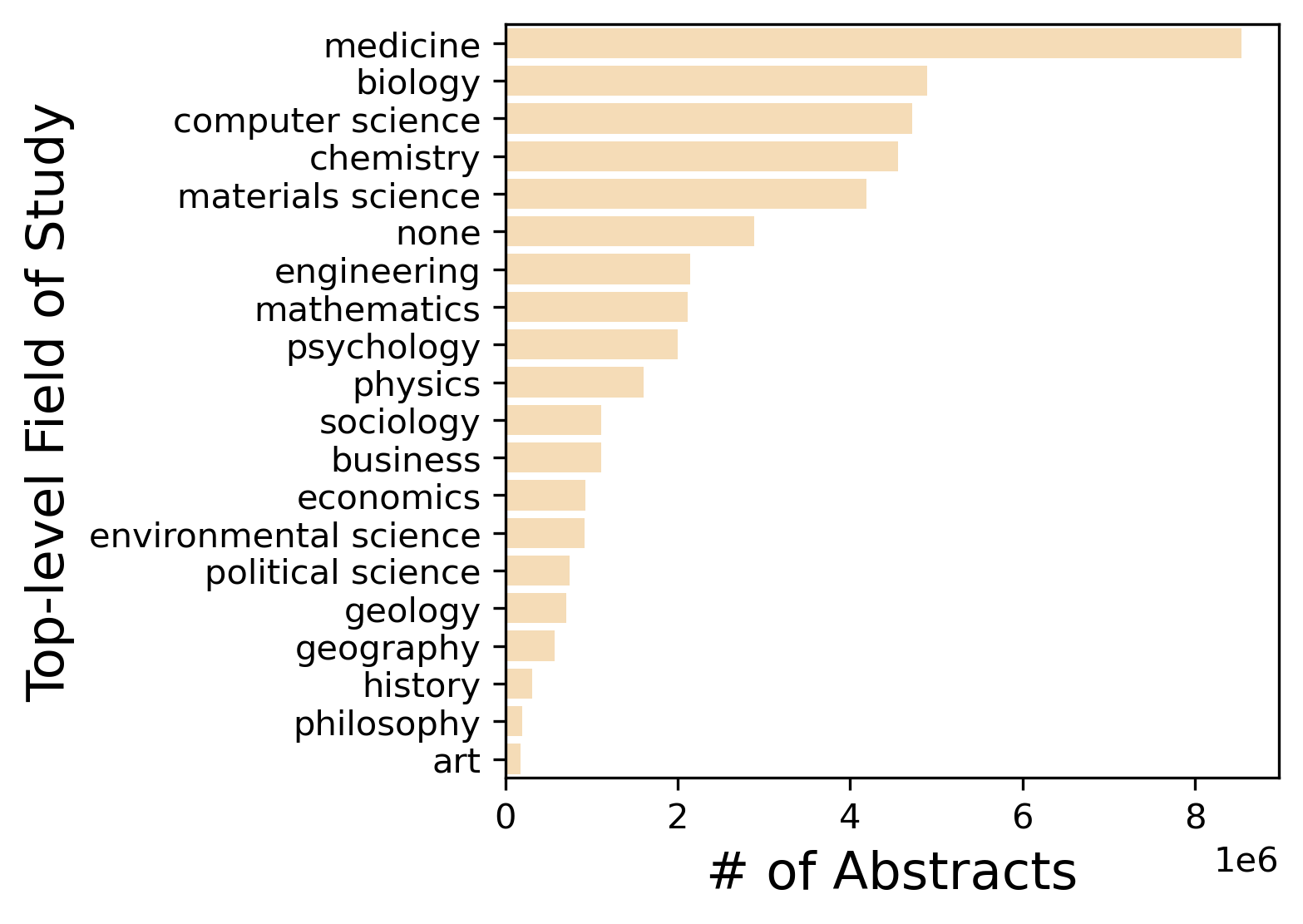}
    \centering
    \caption{The number of abstracts in each top-level MAG field of study (FOS) in S2ORC, sorted from biggest to smallest, before subsampling. Medicine is the most frequent field, and is almost twice as large as the next largest field, which is biology.}
	\label{fig:fos_data_s2orc}
\end{figure}

Figure~\ref{fig:fos_data_s2orc} shows the number of valid abstracts in each top-level MAG field of study \textit{before} subsampling a similar number of abstracts from each subfield. This figure can be compared to Figure~\ref{fig:fos_data} to show how the distribution of fields changed after sampling. The following lists the subfields, or level 1 MAG FOS, used in our study. The same subfield may fall under multiple fields. 

\begin{itemize}
\item \textbf{Art} (6 children): art history, classics, humanities, visual arts, literature, aesthetics.
\item \textbf{Biology} (31 children): computational biology, biochemistry, bioinformatics, cancer research, evolutionary biology, anatomy, molecular biology, pharmacology, immunology, virology, ecology, agronomy, botany, toxicology, food science, microbiology, biological system, agroforestry, biophysics, animal science, paleontology, cell biology, physiology, endocrinology, horticulture, genetics, biotechnology, neuroscience, fishery, zoology, biology (other).
\item \textbf{Business} (10 children): international trade, accounting, risk analysis (engineering), process management, actuarial science, marketing, industrial organization, finance, advertising, business (other).
\item \textbf{Chemistry} (20 children): polymer chemistry, molecular physics, biochemistry, organic chemistry, physical chemistry, chemical physics, nuclear chemistry, medicinal chemistry, photochemistry, combinatorial chemistry, computational chemistry, analytical chemistry, food science, chromatography, mineralogy, inorganic chemistry, crystallography, stereochemistry, environmental chemistry, chemistry (other).
\item \textbf{Computer Science} (32 children): natural language processing, software engineering, theoretical computer science, embedded system, computer security, programming language, data science, computer vision, computer network, human–computer interaction, world wide web, information retrieval, parallel computing, operating system, computer hardware, multimedia, computer graphics (images), library science, real-time computing, artificial intelligence, database, distributed computing, simulation, telecommunications, internet privacy, pattern recognition, machine learning, knowledge management, data mining, speech recognition, algorithm, computer science (other).
\item \textbf{Economics} (28 children): international trade, labour economics, political economy, natural resource economics, industrial organization, monetary economics, economic system, economy, operations management, demographic economics, management, finance, management science, environmental resource management, accounting, agricultural economics, economic growth, actuarial science, financial economics, market economy, socioeconomics, environmental economics, econometrics, law and economics, development economics, public economics, microeconomics, economics (other).
\item \textbf{Engineering} (35 children): engineering ethics, software engineering, control engineering, embedded system, nuclear engineering, reliability engineering, operations research, transport engineering, engineering drawing, biomedical engineering, engineering management, electronic engineering, automotive engineering, forensic engineering, operations management, mechanical engineering, petroleum engineering, process engineering, systems engineering, management science, civil engineering, control theory, simulation, telecommunications, geotechnical engineering, pulp and paper industry, process management, environmental engineering, marine engineering, chemical engineering, manufacturing engineering, waste management, structural engineering, electrical engineering, engineering (other).
\item \textbf{Environmental Science} (7 children): environmental resource management, environmental planning, environmental engineering, agroforestry, soil science, environmental protection, environmental science (other).
\item \textbf{Geography} (7 children): environmental planning, meteorology, archaeology, physical geography, remote sensing, environmental protection, geography (other).
\item \textbf{Geology} (14 children): atmospheric sciences, geochemistry, geomorphology, soil science, hydrology, oceanography, climatology, mineralogy, geotechnical engineering, seismology, petroleum engineering, remote sensing, paleontology, geology (other).
\item \textbf{History} (5 children): art history, classics, ancient history, archaeology, history (other).
\item \textbf{Materials Science} (6 children): polymer chemistry, optoelectronics, composite material, nanotechnology, metallurgy, materials science (other).
\item \textbf{Mathematics} (17 children): geometry, topology, combinatorics, operations research, mathematical optimization, pure mathematics, control theory, discrete mathematics, statistics, algebra, mathematics education, mathematical physics, applied mathematics, econometrics, mathematical analysis, algorithm, mathematics (other).
\item \textbf{Medicine} (45 children): audiology, gerontology, pediatrics, obstetrics, medical physics, urology, radiology, gynecology, dentistry, cancer research, cardiology, veterinary medicine, biomedical engineering, medical education, general surgery, andrology, oncology, dermatology, traditional medicine, orthodontics, anatomy, pharmacology, medical emergency, anesthesia, gastroenterology, immunology, virology, risk analysis (engineering), emergency medicine, surgery, psychiatry, physiology, nursing, endocrinology, clinical psychology, intensive care medicine, physical therapy, nuclear medicine, family medicine, ophthalmology, environmental health, internal medicine, physical medicine and rehabilitation, pathology, medicine (other).
\item \textbf{Philosophy} (6 children): environmental ethics, humanities, epistemology, aesthetics, linguistics, philosophy (other).
\item \textbf{Physics} (24 children): mechanics, atmospheric sciences, molecular physics, astrophysics, acoustics, medical physics, classical mechanics, chemical physics, nuclear physics, optoelectronics, quantum mechanics, theoretical physics, optics, computational physics, particle physics, atomic physics, statistical physics, meteorology, nuclear magnetic resonance, thermodynamics, mathematical physics, astronomy, condensed matter physics, physics (other).
\item \textbf{Political Science} (4 children): public relations, public administration, law, political science (other).
\item \textbf{Psychology} (15 children): mathematics education, cognitive psychology, criminology, clinical psychology, applied psychology, social psychology, communication, pedagogy, psychoanalysis, neuroscience, developmental psychology, psychiatry, psychotherapist, cognitive science, psychology (other).
\item \textbf{Sociology} (11 children): social science, criminology, demography, law and economics, communication, pedagogy, political economy, gender studies, socioeconomics, media studies, sociology (other). 
\end{itemize}

\section{Dataset filtering}\label{appx:filtering}

We perform the following preprocessing steps of S2ORC to create \textsc{Contemporary S2ORC}: 
\begin{itemize}
    \item \textbf{Venue.} We consolidate the \textit{venue} and \textit{journal} keys of each article's metadata. We use whichever label is non-empty, and only a small fraction (0.08\%) of articles with valid abstracts have \textit{venue} and \textit{journal} that differ, in which case we use use the article's \textit{journal}. We handle venue names case insensitively, and also remove tokens in their names that contain numbers to consolidate years and editions. 
    \item \textbf{Time}. Our study focuses on contemporary science, which are abstracts published during 2000-2019. S2ORC contains some abstracts from 2020 and onwards, but dates past 2020 are likely metadata processing errors. We remove $47.6$ million articles outside of this time range. 
    \item \textbf{Valid metadata}. We remove $42.5$ million articles with missing abstracts, titles, or journal and venue labels. 
    \item \textbf{Language.} We remove 77,133 articles from 925 non-English journals or venues, which are those that have less than 80\% of their articles in English, using \citet{lui-baldwin-2012-langid}'s language classifier.
    \item \textbf{Field of study.} Medicine fields dominate S2ORC abstracts. We balance the dataset by taking a sample of 50k articles per subfield. For subfields that are too small to sample or articles that have field-level but no subfield-level labels, we categorize these in an \textsc{other} subfield under their parent field. Since articles can be labeled with multiple FOS, our sample is not perfectly stratified, but prevents large subfields from dominating calculations of the general prevalence of words in English. In total we identify specialized language across 293 subfields that fall under 19 fields (listed in Appendix~\ref{appx:fos}). 
\end{itemize}

\section{Validation details}\label{appx:val}

\begin{figure}[t]
    \includegraphics[width=\columnwidth]{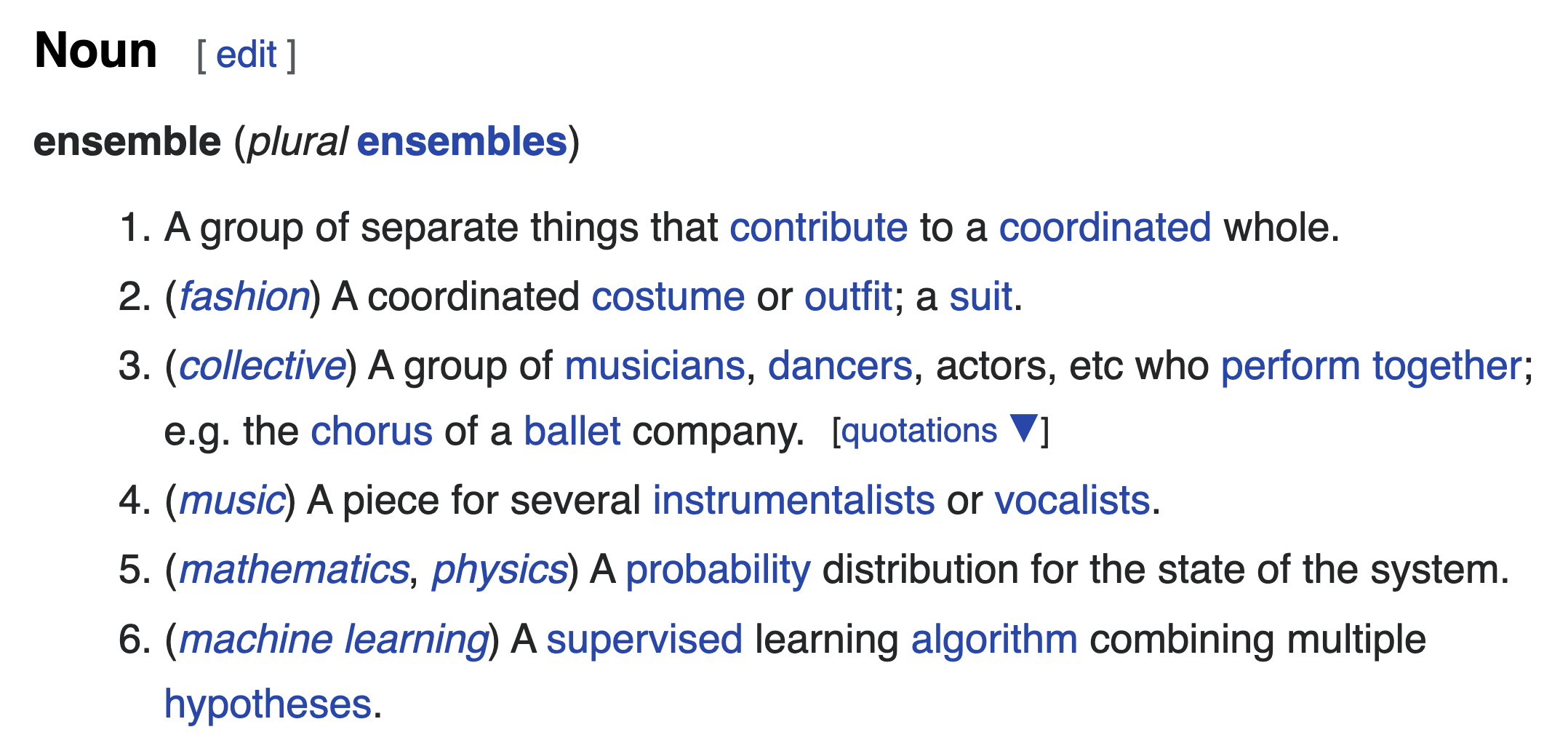}
    \centering
    \caption{An example of a Wiktionary entry for the word \textit{ensemble}. This word has definitions labeled as pertaining to \textit{fashion} (a coordinated outfit), \textit{collective} (a group of performers), \textit{music} (a musical piece), \textit{mathematics \& physics} (a probability distribution), and \textit{machine learning} (a supervised learning algorithm).}
	\label{fig:wiki_example}
\end{figure}

\begin{figure}[t]
    \includegraphics[width=\columnwidth]{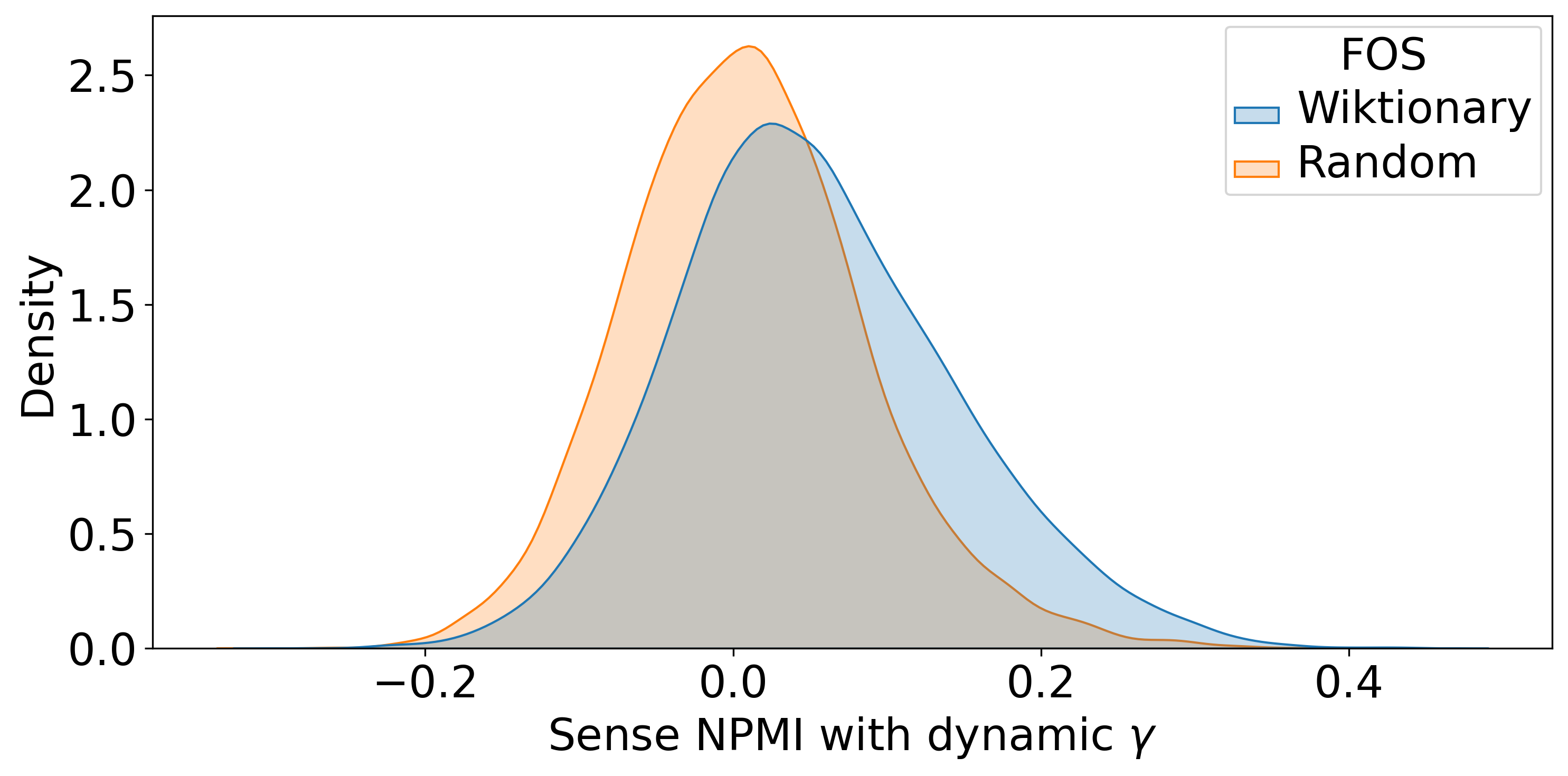}
    \centering
    \caption{The distribution of sense NPMI scores for words in Wiktionary-labeled fields versus random ones. Words labeled as belonging to a subfield by Wiktionary have higher $\mathcal{S}_f(t)$ in that subfield than in a random one (paired $t$-test, $p<0.001$).}
	\label{fig:null_eval}
\end{figure}

Here, we include two additional figures to supplement \S\ref{section:val}.

Figure~\ref{fig:wiki_example} shows a screenshot of a Wiktionary entry for the word \textit{ensemble}, which is overloaded with several labeled definitions.\footnote{\href{https://en.wiktionary.org/wiki/ensemble}{https://en.wiktionary.org/wiki/ensemble}} Some labels, such as \textit{collective}, show grammatical information, while others indicate restricted usage to different fields, dialects, or contexts.\footnote{\href{https://en.wiktionary.org/wiki/Template:label}{https://en.wiktionary.org/wiki/Template:label}} We match these labels to MAG fields and subfields when evaluating recall of words marked as discipline-specific by Wiktionary. 

In the main text, we show that sense NPMI is able to recall more Wiktionary words at the same threshold than type NPMI. In addition, sense NPMI scores are higher in Wiktionary-labeled fields than random ones (Figure~\ref{fig:null_eval}).

\section{Additional experimental details}

\subsection{Cutoff decision}\label{appx:audience}

\begin{figure}[t]
    \includegraphics[width=\columnwidth]{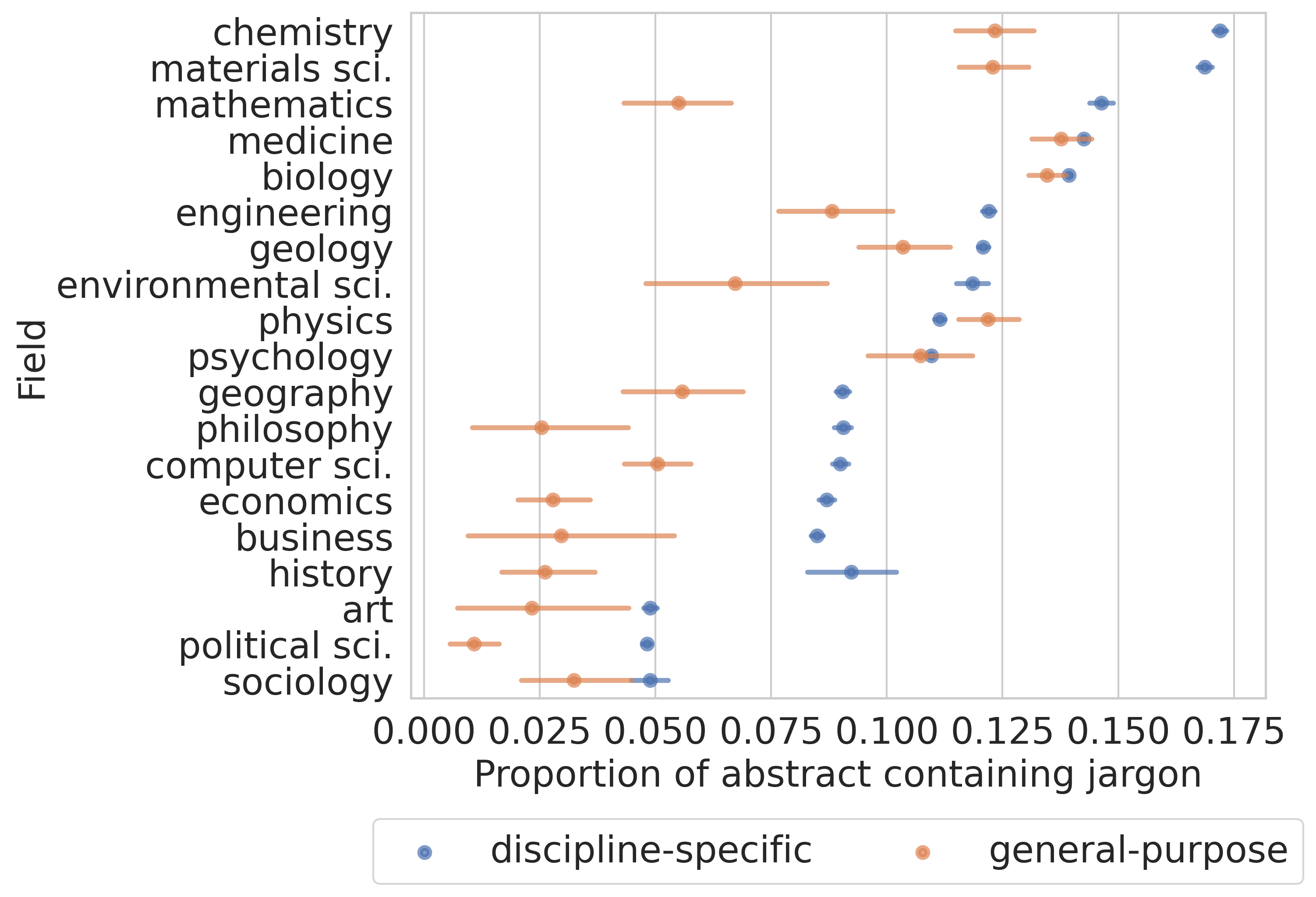}
    \centering
    \caption{Abstracts' average fraction of discipline-specific words and senses in fields that appear in both general-purpose and discipline-focused venues (95\% CI). This plot uses an NPMI cutoff of 0.2.}
	\label{fig:audience_design_appdx}
\end{figure}

\begin{figure}[t]
    \includegraphics[width=0.4\columnwidth]{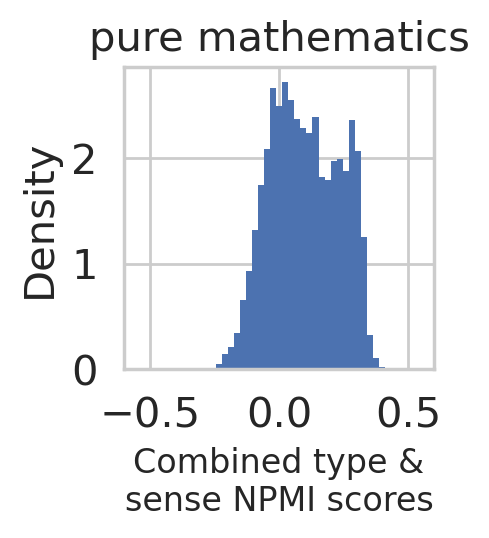}
    \includegraphics[width=0.375\columnwidth]{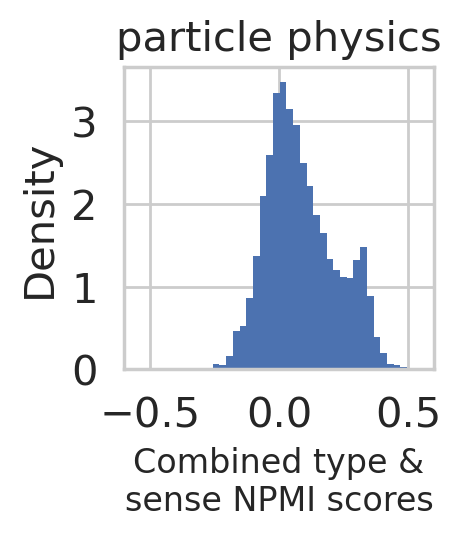}
    \centering
    \caption{Sometimes, NPMI score distributions for subfields are bimodal with a second peak among positive values, especially when a subfield contains large amounts of jargon. The left shows the distribution for pure mathematics, while the right shows particle physics.}
	\label{fig:score_dist}
\end{figure}

We generated Figure~\ref{fig:audience_design} with additional values of the NPMI cutoff $c$, such as $c = 0.2$, and achieve similar conclusions (Figure~\ref{fig:audience_design_appdx}). That is, these results are similar when it comes to which fields tend to adjust their language between general-purpose and discipline-focused venues. In the main text, we usually use $c = 0.1$, as positive NPMI values indicate association, but NPMI values too close to 0 would instead lean towards independence. Though NPMI ranges from -1 to 1, the outputted scores for various subfields tended to range from -0.5 to 0.5, and some include bimodal behavior where the latter peak of the distribution usually occurs after $c = 0.2$ (Figure~\ref{fig:score_dist}). We assume that this latter peak is indicative of jargon. Thus, we experimented with cutoffs that would separate the initial peak around 0 and a secondary peak in the positive NPMI value range, if any. 

\subsection{Scholarly success}\label{appx:success}

\subsubsection{Subfield similarity}

To calculate subfield similarity, we first create a $(N+1) \times (N+1)$ citation matrix, where $N$ is the total number of subfields, and the additional row and column represents articles in unknown subfields. Rows in this matrix represent subfields that are cited, and columns are citing subfields. This matrix is generated using all articles published in S2ORC within the years 2000 and 2019 that have inbound citations. For subfield similarity calculations, we use the rows to represent each subfield. For example, the nearest neighbors via cosine similarity of the row representing \textit{chemical engineering} include \textit{polymer chemistry}, \textit{polymer science}, and \textit{inorganic chemistry}.

\subsubsection{Regressions}

We ran a few statistical tests to determine what regressions to use.

\textbf{Citation counts.} We run both Poisson regressions and negative binomial regressions on citation count data, as these generalized linear models are typically used to model count data. Negative binomial regression is used for data that shows overdispersion, when the variance of the dependent variable exceeds the mean. We calculate the overdispersion ratio $\phi$ of Poisson regressions for each field: $$\phi = \frac{\text{Pearson's } \chi^2}{\text{residual degrees of freedom}}.$$ Since it exceeds 1 for each field's regression, there is overdispersion in our data, and thus we use negative binomial regressions for citation counts. Negative binomial regressions require choosing a constant $\alpha$ which is used to express the variance in terms of the mean. We determine $\alpha$ by inputting the fitted rate vector from the Poisson regression into an auxiliary OLS regression without a constant \cite{cameron2013regression}. The $\alpha$ we obtain from this for each regression is significant for all fields except for Art and Philosophy ($p<0.01$, right-tailed $t$-test).

\textbf{Interdisciplinary impact.} We run ordinary least squares (OLS) regressions for this dependent variable. OLS involves several assumptions: randomly sampled data, linearity, exogeneity, noncollinearity, and homoskedasticity. We check for linearity and exogeneity by comparing residuals and fitted values, non-collinearity by checking that the variance inflation factors of covariates do not exceed 5, and homoskedasticity by running a Breusch-Pagan test \cite{breusch1979test}. We find that we satisfy all assumptions except homoskedasticity. Due to to this, we also run a weighted least squares regression to check the robustness of our OLS results, and achieve similar coefficients.

\end{document}